\documentclass[authoryear,11pt,letterpaper]{elsarticle}
\usepackage[top=0.75in, bottom=0.75in, left=0.75in, right=0.75in]{geometry}

\usepackage{amsmath,amsthm}
\usepackage{amssymb}
\usepackage{bbm}
\usepackage{amsfonts}
\usepackage{graphicx}
\usepackage{subfigure}
\graphicspath{{graphics/}}
\usepackage{algorithm}
\usepackage{algorithmic}
\usepackage{tabularx}
\usepackage{tcolorbox}
\usepackage[customcolors]{hf-tikz}
\hfsetfillcolor{grey!10}
\hfsetbordercolor{white}
\usepackage{tablefootnote}
\usepackage{tabu}
\usepackage{tabularx}
\usepackage{adjustbox}
\tikzset{
  pred/.style={
    set fill color=white,
    set border color=white,
    set disable rounded corners=true,
    above left offset={-0.1,0.3},
    below right offset={0.75,-0.65},
  },
  obs/.style={
    set fill color=grey!10,
    set border color=white,
    set disable rounded corners=true,
  }
}

\usepackage{color,xcolor}
\definecolor{darkblue}{rgb}{0.0,0.5,0.5}
\definecolor{blue}{rgb}{0.0,0.5,0.68}
\usepackage[colorlinks]{hyperref}
\hypersetup{colorlinks,breaklinks,linkcolor=darkblue,urlcolor=darkblue,anchorcolor=darkblue,citecolor=darkblue}
\usepackage{booktabs,caption}
\usepackage{multirow}
\usepackage{lscape}
\usepackage{epstopdf}
\usepackage{lineno}
\usepackage{subfigure}
\usepackage{mathrsfs,MnSymbol}
\usepackage{longtable}
\usepackage{microtype}
\usepackage{lineno}
\usepackage{colortbl}

\definecolor{Gray}{gray}{0.9}
\newdefinition{definition}{Definition}

\journal{ }
\makeatletter
\def\ps@pprintTitle{%
   \let\@oddhead\@empty
   \let\@evenhead\@empty
   \let\@oddfoot\@empty
   \let\@evenfoot\@oddfoot
}
\makeatother

\begin{document}

\begin{frontmatter}



\title{Advancing Transportation Mode Share Analysis with Built Environment: Deep Hybrid Models with Urban Road Network}



\author[addmit]{Dingyi Zhuang}
\ead{dingyi@mit.edu}

\author[addmit]{Qingyi Wang}
\ead{qingyiw@mit.edu}

\author[addmit]{Yunhan Zheng}
\ead{yunhan@mit.edu}

\author[addmit]{Xiaotong Guo}
\ead{xtguo@mit.edu}

\author[adduf,addmit]{Shenhao Wang\corref{cor1}}
\ead{shenhaowang@ufl.edu}

\author[addneu]{Haris N. Koutsopoulos}
\ead{h.koutsopoulos@northeastern.edu}

\author[addmit]{Jinhua Zhao}
\ead{jinhua@mit.edu}

\address[addmit]{Department of Civil and Environmental Engineering, Massachusetts Institute of Technology, Cambridge, MA, USA}
\address[addneu]{Department of Civil and Environmental Engineering, Northeastern University, 360 Huntington Avenue, Boston, MA, USA}
\address[adduf]{Department of Urban and Regional Planning, University of Florida, Gainesville, FL, USA}
\cortext[cor1]{Corresponding author. Address: Arch building, \#434, 1480 Inner Rd, Gainesville, Florida 32611, USA}

\begin{abstract}
Transportation mode share analysis is important to various real-world transportation tasks as it helps researchers understand the travel behaviors and choices of passengers. 
A typical example is the prediction of communities' travel mode share by accounting for their sociodemographics like age, income, etc., and travel modes' attributes (e.g. travel cost and time).
However, there exist only limited efforts in integrating the structure of the urban built environment, e.g., road networks, into the mode share models to capture the impacts of the built environment. 
This task usually requires manual feature engineering or prior knowledge of the urban design features. 
In this study, we propose deep hybrid models (DHM), which directly combine road networks and sociodemographic features as inputs for travel mode share analysis. 
Using graph embedding (GE) techniques, we enhance travel demand models with a more powerful representation of urban structures. 
In experiments of mode share prediction in Chicago, results demonstrate that DHM can provide valuable spatial insights into the sociodemographic structure, improving the performance of travel demand models in estimating different mode shares at the city level. 
Specifically,  DHM improves the results by more than 20\% while retaining the interpretation power of the choice models, demonstrating its superiority in interpretability, prediction accuracy, and geographical insights. 
\end{abstract}

\end{frontmatter}


\section{Introduction}

Mode share analysis in transportation quantifies the usage distribution among different transport modes, like cars, public transit, and bicycles, within a certain area. 
It usually takes various factors into account, like age, income, travel time, cost, etc., and applies travel demand models to analyze the mode share.
It aims to help researchers in planning and policy-making by highlighting trends and priorities for sustainable and efficient mobility solutions. \citep{li2017bicycle,bucsky2020modal,ben1985discrete,ben1999discrete}

In recent decades, the world has experienced a rapid surge in urbanization, resulting in significant changes to the built environment. 
This change has a direct impact on people's daily commuting behavior, mode share, and trip purpose, which should be properly incorporated into mode share analysis. 
For instance, an increased number of rail/bus stops and a less extensive road network can encourage people to use public transportation for commuting purposes \citep{strano2017scaling,kalapala2006scale}. 
Therefore, comprehending the built environment is essential to studying the infrastructure-side influence on travel demand. 

As a consequence, mode share analysis that considers the built environment as input is becoming increasingly popular in the transportation field. 
They provide a comprehensive view of the relationship between the physical environment and travel behavior, taking into account factors like land use mix, road network characteristics, and public transportation accessibility, which can affect travel behavior by influencing the availability and accessibility of transportation options \citep{wang2020deep,wang2020deep2,koppa2022deep}. 
These models offer a more nuanced understanding of the relationship between the physical environment and travel behavior than traditional utility-based models that only include trip-based characteristics.

However, integrating the built environment into the analysis is a complex and demanding task, which requires improvement through the technical framework.
Existing work that considers the built environment requires prior knowledge and manual feature engineering to extract useful features for travel demand models to analyze the mode share.
This is because the built environment data are unstructured, which provide non-tabular and non-quantitative information \citep{lam2002combined,garling2009travel}.
Take urban road network structure as the example, researchers need to manually create accessibility, design, diversity, and other features to represent its information in order to study their effect on mode share analysis \citep{cervero1997travel, marshall2010effect,bettencourt2007growth,arcaute2015constructing,xu2020deconstructing}.
These process require prior knowledge, additional data (such as land use data), and a well-designed feature set, which can make it difficult to extend the knowledge to other cities.

Given these challenges, we propose the framework of \textit{Deep Hybrid Models (DHM)} which integrates deep learning (DL) and hybrid models to examine the built environment impacts of urban road networks \citep{wang2023deep}. 
We utilize the advantages of graph embedding (GE) techniques, \textit{Node2Vec}, to directly learn the latent variables from urban road networks as the key intermediate step to build travel demand models.
We develop a framework that can represent the features of the road network without prior knowledge and reflect the interplay between the road network structure, functionality, and urban demographics.
We open the gate of analyzing the impactful built environment factors by starting from the urban road network structure, due to road networks' high accessibility to all transportation modes. 
Meanwhile, it is challenging to design measurements and extract meaningful information due to its unstructured data formats that cannot be directly processed by classic analytical models \citep{wang2012understanding,zhan2017dynamics,li2015percolation,saberi2020simple}. 

The contributions of this paper can be summarized into threefold:

\begin{itemize}
\item We address the use of urban road structures as inputs for travel demand models, an area that has been less studied before in mode share analysis. 
It opens the gates to potentially integrate other types of unstructured built environment data.
\item Our DHM framework, combining deep learning and (hybrid) choice models, significantly enhances the regression power of travel demand models and retains interpretability.
\item We interpret the impact of urban road networks, GE techniques, and patterns of Graph Embedding Representations (GER), providing geographical and urban planning insights for analyzing travel mode share and community structure.
\end{itemize}

The paper is structured as follows: in Section \ref{sec:review} we review the related works; in Section \ref{sec:method} we introduce our GE methodology \textit{Node2Vec} and the proof of concept of the DHM; in Section \ref{sec:exp} we implement a case study in Chicago to highlight the regression performance; in Section \ref{sec:interpretation} we give visual interpretations of GER and its physical meaning; lastly, we conclude and discuss the future work in Section \ref{sec:conclusion}.

\section{Literature Review}
\label{sec:review}
\subsection{Mode Share Analysis in Transportation Engineering}

Mode share analysis is a critical component of transportation engineering, profoundly influencing infrastructure planning, policy formulation, and system optimization. 
Given a specific region or area, it focuses on understanding the aggregated portions of people's preferences across different transportation modes—such as private cars, public transport, taxis, or non-motorized options—and the consequent impact on urban planning and sustainability. 
Traditionally, mode share analysis has relied on demand models like choice models (CMs) to predict transportation mode shares, based on a mix of observable and latent factors, emphasizing utility maximization \citep{ben1996travel,cantarella2005multilayer,train2009discrete,koppa2022deep,small2007economics}.

A remark of these traditional analyses was the emphasis on sociodemographic data. 
Factors such as age, income, employment status, and household size were considered vital in shaping travel behavior \citep{ben1985discrete,ben1995discrete,salas2022systematic}. 
The methodology revolved around the hypothesis that individuals with different socioeconomic backgrounds would display discernibly varied transportation mode preferences \citep{abou2010model,hasnine2018dynamics,small1998demand,smilkov2017smoothgrad}. 
For instance, communities with higher income levels might opt for private cars, while those with limited financial resources might lean towards public transportation or non-motorized modes. 
This reliance on structured sociodemographic data enabled researchers and planners to derive patterns and insights, albeit within the constraints of the data's granularity and scope.

While structured sociodemographic data has been invaluable, there's an increasing recognition of the role of unstructured data in refining mode share predictions \citep{yang2023impact,snellen2002urban,zhang2004role}. 
The unstructured data are the data type that is non-tabular and non-quantitative information to analyze directly, such as word-processing text documents, images and video files, and so on \citep{blumberg2003problem}.
Among the unstructured data, road network data provides critical insights into the physical environment of transportation. 
By analyzing road network topologies, junction densities, and connectivity indices, researchers can better understand factors such as travel times, accessibility, and even the appeal of non-motorized modes like walking or cycling in particular regions \citep{cooper2017using,scheepers2016perceived}.

However, despite the depth and nuance in mode share analysis provided by traditional CMs, their inherent design has often been criticized for offering limited flexibility in adapting to newer, unconventional data streams, like unstructured data.
This necessitates an evolved modeling framework that synergizes the logical foundation of CMs with the adaptive prowess of advanced computational techniques.

\subsection{Hybrid Models in Transportation Engineering}
Bridging the divide between the traditional CMs and the adaptability required in various urban data formats, hybrid models emerge as the forerunners of next-gen transportation analysis. 
These models, as the name suggests, blend the strengths of two distinct analytical realms: CMs and machine learning (ML) or DL techniques. 
While the former brings structure, hypothesis testing, and behavioral insights, the latter adds the versatility to handle diverse data forms and the capability to discern patterns in high-dimensional spaces. 
Such models show potential in yielding more precise travel behavior patterns \citep{de2011modelling,small2007economics,zheng2023fairness}. 
Traditional CMs, being theory-driven, struggle to capture multifaceted relationships. 
In contrast, ML and DL represent rapidly advancing domains within artificial intelligence capable of modeling intricate relationships in data. 
This ability has sparked growing interest in their deployment in transportation planning \citep{Wang2020DeepInterpretation,Wang2020DeepFunctions}.

The integration of ML/DL and choice models is defined as hybrid models, which aims to leverage the advantages of both models. 
It can capture rational passenger behavior and complex relationships among travel demand, socioeconomic factors, and transportation network characteristics. 
According to \citet{van2021choice}, ML and DL can assist in finding utility functions and identifying systematic and random heterogeneity based on the original choice model. 
Hybrid models maintain the interpretability of the choice model while overcoming the limitations of traditional methods using ML techniques. For instance, \citet{han2022neural} used a neural network to learn the representation of taste heterogeneity while keeping the utility function the same as the random utility function with heterogeneous taste. 
Hybrid models are particularly useful in mode share analysis, which predicts the portions of travelers choosing a particular mode of transportation. 
The combination of traditional CMs and ML algorithms in hybrid choice models leads to more accurate predictions, particularly in complex and rapidly changing environments.

Latent variables are also an essential concept in choice models as they capture psychometric features like individual preferences and traits, which can explain the hidden correlations among different observed variables \citep{cho2016latent,greene2003latent}. 
The high-dimensional space in which these latent variables reside is referred to as the latent space. 
Structural equation models with indicators collected in surveys can estimate latent variables, helping to formulate preference heterogeneity among the population and explain behavior differences across population groups \citep{Vij2016}. 
By integrating the concepts of latent variables and random utility theory, hybrid choice models are developed to describe the joint impacts of latent variables and utilities \citep{Walker2002, Ben-Akiva2002HybridChallenges}. 
The concept of the latent variable as well as latent space is similar to representation learning (e.g., embedding) in deep learning to reflect the underlying correlation among the input variables as they both create dimensions to formulate the observed data \cite{wang2023deep,Thorhauge2019}. 
However, current CMs do not consider other factors such as infrastructure impacts. Therefore, there is a need to further enhance both hybrid models and latent space concepts to address such limitations.

\subsection{Graph Embedding of Urban Road Network Topology}

In order to transform the road network structures into the latent space, we introduce the Graph embedding technique.
It is a machine learning method that can transform graph-structured data into a low-dimensional Euclidean space. 
By using vector representations of nodes (i.e. GER), these techniques can facilitate various downstream tasks, including node classification, link prediction, and clustering. 
The application of GE has shown promising performance improvements in a range of areas, such as social network embedding, human mobility analysis, and drug discovery, and so on \citep{cheng2019network,jalili2017link,lerique2020joint,ren2014predicting,teney2017graph}. 
In particular, GE has been widely applied in traffic flow prediction for transportation planning. 
By embedding sensor networks and historical traffic data into a high-dimensional space, machine learning algorithms can learn patterns and correlations between different road segments and traffic volumes \citep{wu2021inductive, zhuang2022uncertainty}. 
This can be used to predict traffic flow in real-time, which is valuable for traffic management, safety control, and congestion reduction \citep{xu2019road,wu2021spatial,zhuang2020compound,liu2022universal,xu2021understanding,jiang2023uncertainty,wang2023uncertainty,wu2020learning}.

Various GE techniques have been proposed, including matrix factorization, random walk, and deep learning based methods. 
Matrix factorization-based methods, such as graph factorization and matrix decomposition, map nodes to vectors by approximating the adjacency matrix of the graph \citep{qiu2018network,liu2019general,rozemberczki2019gemsec}. 
Random walk-based methods, such as \textit{DeepWalk} and \textit{Node2Vec}, map nodes to vectors by simulating random walks on the graph and learning a representation based on the transitions between nodes \citep{grover2016node2vec,perozzi2014deepwalk}. 
Deep learning-based methods, such as graph convolutional networks (GCN) and graph attention networks (GAT), map nodes to vectors by incorporating the graph structure and node features into a neural network architecture \citep{cai2018comprehensive,goyal2018graph}.

The built environment, especially the road network, is an important factor that impacts travel behavior and is well suited for GE techniques.
It is often measured by the density, diversity, and design, known as the "3Ds" \citep{Ewing2010TravelEnvironment, Yin2020ExploringMatter} in previous literatures. 
However, 3Ds are difficult to measure, due the cost of time and manpower, making it challenging to draw generalizable conclusions. 
GE can be introduced to directly capture the geometric, topological, and semantic information of roads, including location, length, direction, and category, as well as relationships between different roads, such as intersections, connections, and proximity. 
The learned embeddings can be used to support various urban planning and management tasks, such as traffic prediction, route optimization, and emergency response. 
For example, \citet{xue2022quantifying} applied \textit{Node2Vec} on road networks across multiple cities worldwide to study spatial homogeneity patterns and \citet{xu2020ge} applied \textit{DeepWalk} on road networks to assist in predicting the traffic states.

CMs and hybrid models are limited by the representation power of unstructured data. 
Therefore, using advanced GE methodology to replace original handcrafted features can be inspiring. 
By formulating urban road networks in a way that can fit choice models, it may be possible to improve the accuracy and effectiveness of such models.

\section{Methodology}
\label{sec:method}

\subsection{Problem Description}
Our proposed framework aims to integrate deep learning and hybrid choice modeling to improve the built environment impacts in mode share analysis.
Thus, the problem formulation consists of two major components: (1) how to represent of urban road networks using GE, and (2) how to construct a mode share model based on demand and sociodemographic data and learned GERs.

To abstractly represent a city's road network, we treat intersections as nodes and road segments as edges, where the sets of nodes and edges are represented as $\mathcal{V}$ and $\mathcal{E}$, respectively. 
The adjacency matrix $A$, encapsulating road topology and travel distance, is employed to construct a road network graph, $\mathcal{G} = (\mathcal{V},\mathcal{E},A)$. 
The first component of our problem involves deriving a GER of the road network, expressed as $\mathcal{R}(K^*) = Embedding(\mathcal{G})$. 
Here, $K^*$ represents the dimension of our tailored GER, and $Embedding$ symbolizes the GE technique applied. 
For concise notation, we utilize $\mathcal{R} \in \mathbb{R}^{N\times K^*}$ for the node embedding results $\mathcal{R}(K^*)$. 
Therefore, the representation of an individual node $u$ should be $\mathcal{R}_{u} \in \mathbb{R}^{1\times K^*}$. 
These embeddings are treated as latent variables within the hybrid choice model framework, as they are generated from observed variables and reflect inherent correlations. 
Notice that in the classic mode share analysis, we might use features $x\in \mathbb{R}^{N\times K}$ as the inputs, which is usually collected from both supply and demand side of the traffic service. 
Note that $K$ denotes the size of the sociodemographic features.

The second component involves constructing a travel demand model to predict the mode shares based on $\mathcal{R}$ or $x$. 
We will demonstrate it in the manner of CM. In a conventional CM task, each of the $N$ census tracts has the mode set $C_n$. 
The mode set $C_n$ encompasses the driving, public transit, walking, or other options, aggregating all individual surveys from the $n$-th census tract. 
In earlier CM studies, the observed data for census tract $n$ includes sociodemographic attributes of each alternative $x_{in}, \forall i \in C_n$, and the mode share $y_n$. 
The fraction of census tract $n$ opting for alternative $i$ is $P(y_n=i | x_{jn}, \forall j \in C_{n})$. 
In the simplest form, we learn that the systematic utility $V_{in}$ of alternative $i$ for a census tract $n$ can be defined as a linear combination of $K$ attributes (Equation \ref{eq:origin}), where $\beta_{ki}$ signifies the parameters for attribute $x_{ki}$ and $\beta_{i0}$ denotes the alternative-specific constant for alternative $i$.

\begin{equation}
\label{eq:origin}
V_{in} = \beta_{i0} + \sum_{k=1}^{K} \beta_{ik} x_{ikn}.
\end{equation}

For our road network-related choice task, it suffices to replace the sociodemographic attributes $x_{ki}$ with our GER. 
Consequently, we can readily restate the equation as Equation \ref{eq:ger_lm}:

\begin{equation}
\label{eq:ger_lm}
V_{in} = \beta_{i0} + \sum_{k=1}^{K^*} \beta_{ik} \mathcal{R}_{ikn}.
\end{equation}

After that, we applied predictive travel demand models, like multinomial logit (MNL) choice model or machine learning regressors, to estimate The probability that an individual from census tract $n$ selects alternative $i$ as:

\begin{equation}
    P(y_n = i | V_{in}) = g(V_{in}) =  \frac{e^{V_{in}}}{\sum_{j \in C_n} e^{V_{jn}}}.
\label{eq:softmax}
\end{equation}

Since we only care about the portions of the travel modes, we keep the obtained results in Equation \ref{eq:softmax} without discretizing them \citep{Ben-Akiva2002IntegrationModels,Ben-Akiva1985DiscreteDemand}. 
It is worth noting that $g$ could be replaced by other choice and machine learning models could be used in lieu of the linear utility form, apart from MNL model. 
However, the central aim of this paper is to highlight the potential of integrating unstructured data with traditional travel demand modeling. 
Therefore, we use the simplest choice modeling form to underline the advantages of employing GE.

\subsection{Graph Embedding Method: \textit{Node2Vec}}

Our initial task is to learn the GER to serve as input for the travel demand model. 
For this task, we employ the \textit{Node2Vec} embedding model, as we found it more suitable than GNN alternatives like graph auto-encoders (GAEs) \citep{grover2016node2vec,pan2018adversarially,kipf2016variational,wang2017mgae}. 
While GAE, which includes encoder and decoder blocks, aims to reconstruct the graph adjacency, its application requires the definition of node features as inputs (e.g. traffic volume and speed), which necessitates specific design considerations. 
Conversely, the \textit{Node2Vec} model, based on random walk sampling, requires only the adjacency matrix $A$ as input and generates the embedding for each node as output. 
This approach is not only more intuitive and flexible but also efficient, as the edge transition probability, once calculated, remains fixed for a given road network of a city. 
This consistency provides transferability and exploratory power when applied to different cities.

 The process of implementing \textit{Node2Vec} encompasses three stages: computation of edge transition probabilities, random walk sampling, and embedding calculation. 
 The neighborhood of a node $u$, represented as $N_S(u)$, is determined based on the neighborhood sampling strategy $S$. 
 We initiate a $2^{nd}$ order random walk process by selecting a source node $u$ and simulating a random walk of customized fixed length $l$. 
 The $i$th node in the walk, $c_i$, is generated based on the following distribution:

\begin{equation}
    P(c_i = x | c_{i-1} = v)=\left\{
    \begin{array}{lc}
    \frac{\pi_{vx}}{Z} & \text{if}\quad (v,x)\in \mathcal{E} \\
    0 & \text{otherwise}
    \end{array}
    \right.,
\end{equation}
\noindent where $\pi_{vx}$ is the unnormalized transition probability between nodes $v$ and $x$, and $Z$ is the normalizing constant. 
$\pi_{vx}$ is inversely proportional to travel distance, resulting in more frequent visits to proximate roads. \citet{grover2016node2vec} proposed a mixed Breadth-First Search (BFS) and Depth-First Search (DFS) sampling strategy with two parameters; $p$ and $q$, standing for the return and in-out parameters respectively. 
These parameters control the probability of immediately returning to a node just visited during the walk and the likelihood of the walk visiting nodes further away from a source node $u$.

Assume a random walk has just traversed edge $(t,v)$ and arrived at node $v$. 
The walk then evaluates the transition probability $\pi_{vx}$ on edges $(v,x)$ from $v$. 
The unnormalized transition probability can be reformulated as $\pi_{vx} = \alpha_{pq}(t,x)\cdot w_{vx}$, where $w_{vx}$ is the edge weight (i.e., road network length), and $\alpha_{pq}$ is the search bias, a function of $p$ and $q$ defined as:

\begin{equation}
    \alpha_{pq}(t,x)=\left\{
    \begin{array}{lc}
    \frac{1}{p} & \text{if}\quad d_{tx}=0 \\
    1 & \text{if}\quad d_{tx}=1 \\
    \frac{1}{q} & \text{if}\quad d_{tx}=2 
    \end{array}
    \right.,
\end{equation}

\noindent where $d_{tx}$ is the shortest path distance between nodes $t$ and $x$ and must be one of ${0,1,2}$ so that two parameters are sufficient and necessary to guide the random walk. 
Parameters $p$ and $q$ control the exploration speed of the random walk. 
The random walk process provides us with the sampling strategy $S$ and the likelihood of visiting the neighboring node $u$. 
We aim to make our node embedding as similar as possible to the embeddings of nodes in its neighborhood. 
Thus, we optimize the objective function that maximizes the log-probability of observing the neighborhood $N_S(u)$ of node $u$, given as:

\begin{equation}
\max_{{\mathcal{R}{u}}} \sum_{u\in \mathcal{V}}\log Pr(N_S(u)|\mathcal{R}_{u}).
\label{eq:obj_origin}
\end{equation}

To render the optimization problem tractable, we introduce two assumptions:

\begin{itemize}
\item Conditional Independence: The log-sum operation is based on the assumption of independence of the likelihood of observing a neighborhood from observing any other neighborhood node, i.e., $Pr(N_S(u)|\mathcal{R}_{u}) = \prod_{v_i \in N_S(u)}Pr(v_i|\mathcal{R}_u)$.
\item Symmetry in Feature Space: A source node and its neighboring node in $N_S(u)$ have a symmetric effect in the feature space. 
In our context, as our road network includes only the road connections between intersections, we disregard the directions of the road. 
The conditional likelihood of each source-neighborhood node pair can be parameterized as a softmax function: $Pr(v_i|\mathcal{R}u) = \frac{\exp(\mathcal{R}{v_i} \cdot \mathcal{R}u ) }{ \sum{v\in \mathcal{V}} \exp(\mathcal{R}{v} \cdot \mathcal{R}_u) }$, where $v_i \in N_S(u)$.
\end{itemize}

Given these assumptions, Equation \ref{eq:obj_origin} can be simplified into its final form:

\begin{equation}
\max_{{\mathcal{R}{u}}} \sum_{u\in \mathcal{V}} [-\log Z + \sum_{v_i \in N_S(u)} \mathcal{R}_{v_i} \cdot \mathcal{R}_u ],
\label{eq:obj_final}
\end{equation}

\begin{table}[t]
    \centering
    \small
    \begin{tabular}{c|c}
    \hline
       Parameter  & Value \\
    \hline
        $K^*$ (Size of GER)  & 128\\
        \# of walks for each node & 10\\
        Walk length & 20\\
        \# of negative samples to use for each positive sample  & 1\\
        Return parameter $p$ & 1\\
        In-out parameter $q$ & 1\\
        \# of epochs to run & 100\\
        Learning rate & 0.01\\
    \hline
    \end{tabular}
    \caption{Parameter tables of our implemented \textit{Node2Vec} model.}
    \label{tab:params_node2vec}
\end{table}

In the above equation, the per-node partition function, $Z_u = \sum_{v\in \mathcal{V}} \exp(\mathcal{R}_v \cdot \mathcal{R}_u) $, can be computationally expensive for large networks. 
To improve efficiency, we employ negative sampling, which reduces the number of training examples needed by randomly sampling negative examples (i.e., node pairs not connected by an edge) to train alongside positive examples (i.e., node pairs connected by an edge). 
This strategy allows the model to learn to distinguish between positive and negative examples more efficiently, without the need to generate all possible negative examples. 
Equation \ref{eq:obj_final} is learned via a two-layer neural network and stochastic gradient ascent applied to the model parameters, yielding the embedding $\mathcal{R}(k)$.

Note that many GE techniques, including \textit{Node2Vec}, primarily focus on node-level representation, which means they generate the embedding vector $\mathcal{R}_u$ for node $u$ instead of the representation of the census tract embedding $\mathcal{R}_n$ \citep{hamilton2017inductive}. 
To convert node-level embedding into census tract (i.e., subgraph) level, we need an additional aggregation step, known as "readout" \citep{xu2018powerful}. 
For each embedding vector dimension, the readout function is defined as:

\begin{equation}
\mathcal{R}_{n,k} = \frac{1}{|\mathcal{G}n|} \sum_{u\in \mathcal{G}_n} \mathcal{R}_{u,k},
\label{eq:readout}
\end{equation}

where $\mathcal{G}_n$ is the subgraph comprising all nodes and their interconnected edges within the census tract $n$, with $|\mathcal{G}_n|$ denoting the number of nodes within $\mathcal{G}_n$. 
In our implementation, the parameters for \textit{Node2Vec} are shown in Table \ref{tab:params_node2vec}.

\subsection{Deep Hybrid Model}

\begin{figure}[htbp]
\centering
\includegraphics[width = 0.7\textwidth]{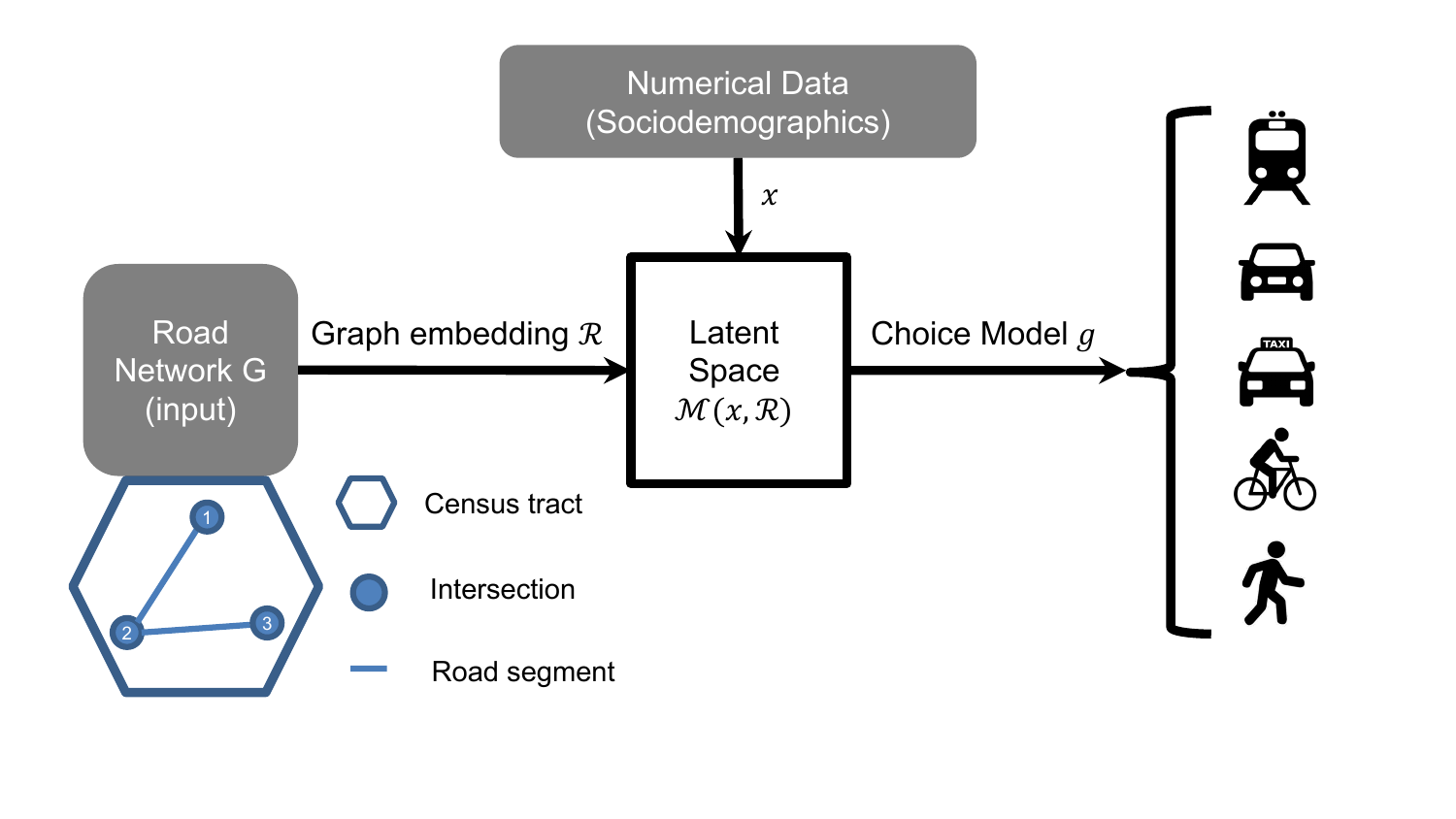}
\caption{A proof-of-concept demonstration of DHM applied to an urban road network for mode share analysis.}
\label{fig:dhm}
\end{figure}

We propose an innovative and comprehensive framework known as the Deep Hybrid Models (DHMs), seeking to unify the principles of hybrid choice modeling and the power of the machine and deep learning to accurately represent unstructured data \citep{wang2023deep}. 
The DHM fuses different domains and data types. 
As depicted in Figure \ref{fig:dhm}, we provide a conceptual demonstration of how DHM could be employed in the context of an urban road network.

For a more formal representation following the previous notations, DHM can be mathematically defined as follows:

\begin{equation}
P(y_n = i | V_{n}) = g(V_{n}) = g(z_{n}) = g(\mathcal{M}(x_n, \mathcal{R}_{n})),
\label{eq:dhm_master}
\end{equation}
In the above equation, for the $n$-th census tract, $x_{n}$ signifies numerical inputs, such as sociodemographics, and $\mathcal{R}_{n}$ represents graph-embedded variables. 
The DHM framework is characterized by two main components brought together as $\mathcal{M}(x_{n}, \mathcal{R}_{n})$, while the function $g(\cdot)$ acts as a prediction function, like MNL, estimating the output probability. 
Therefore, the DHM framework essentially comprises a \textit{mixing operator}, denoted as $\mathcal{M(\cdot)}$, and a \textit{behavioral predictor}, denoted as $g(\cdot)$. 
Notably, the mixing operator could be a simple fusion of the $x_n$ and $\mathcal{R}_{n}$, or even just a segment of it. 
The behavioral predictor adopts a generalized linear form: $g(z_n) = \sigma(\beta'z_n)$, where $\sigma(\cdot)$ symbolizes the link function and $\beta'z_n$ is a linear transformation of the latent variables ($z_n = \mathcal{M}(x_n, \mathcal{R}_n)$). 
It is worth mentioning that we have intentionally simplified $g(\cdot)$ for the sake of focusing on the mixing operator. 
However, $g(\cdot)$ possesses the flexibility to handle a diverse range of output categories, such as single variable outputs, soft choice probabilities, and discrete choices.

In a similar vein to the hybrid choice model, the DHM leverages a latent variable $z_n$ within a latent space to encapsulate complex alternative information — in this case, unstructured road network data. 
Our proposed GE architecture adeptly transforms the road network structures into a high-dimensional latent space. 
The concept of a latent space and latent variables remains relevant and advantageous. DHM, in fact, enhances this concept with its versatility; the latent space can also incorporate and interact with supplementary data such as sociodemographics. 
Moreover, the variables residing in the latent space can directly serve as inputs for travel demand models, enhancing the predictive power of the model.

Another key feature of DHM is its adaptability. 
Both the input and output components of DHM can be configured to cater to different tasks, making it a highly flexible solution that can be employed across various problem domains. 
Further extending its versatility, DHM can process other forms of unstructured data, including images and texts, by substituting the GNN encoder with other suitable embedding techniques, like Convolutional Neural Networks (CNNs), Recurrent Neural Networks (RNNs), or Transformers. 
This substitution capability makes DHM a scalable and robust framework, able to handle a wide array of data types and tasks.

\section{Experiments}
\label{sec:exp}
\subsection{Data Resources}
We use Chicago to demonstrate the proposed model and include multiple data sources. Notice that our research scope is at census tract level, which means our inputs $x_n$ and $\mathcal{R}$ come from the sociodemographics and road network structure from the census tract.

\subsubsection{Road Network Data Description}
\begin{figure}[htbp]
    \centering
    \includegraphics[width=0.6\textwidth]{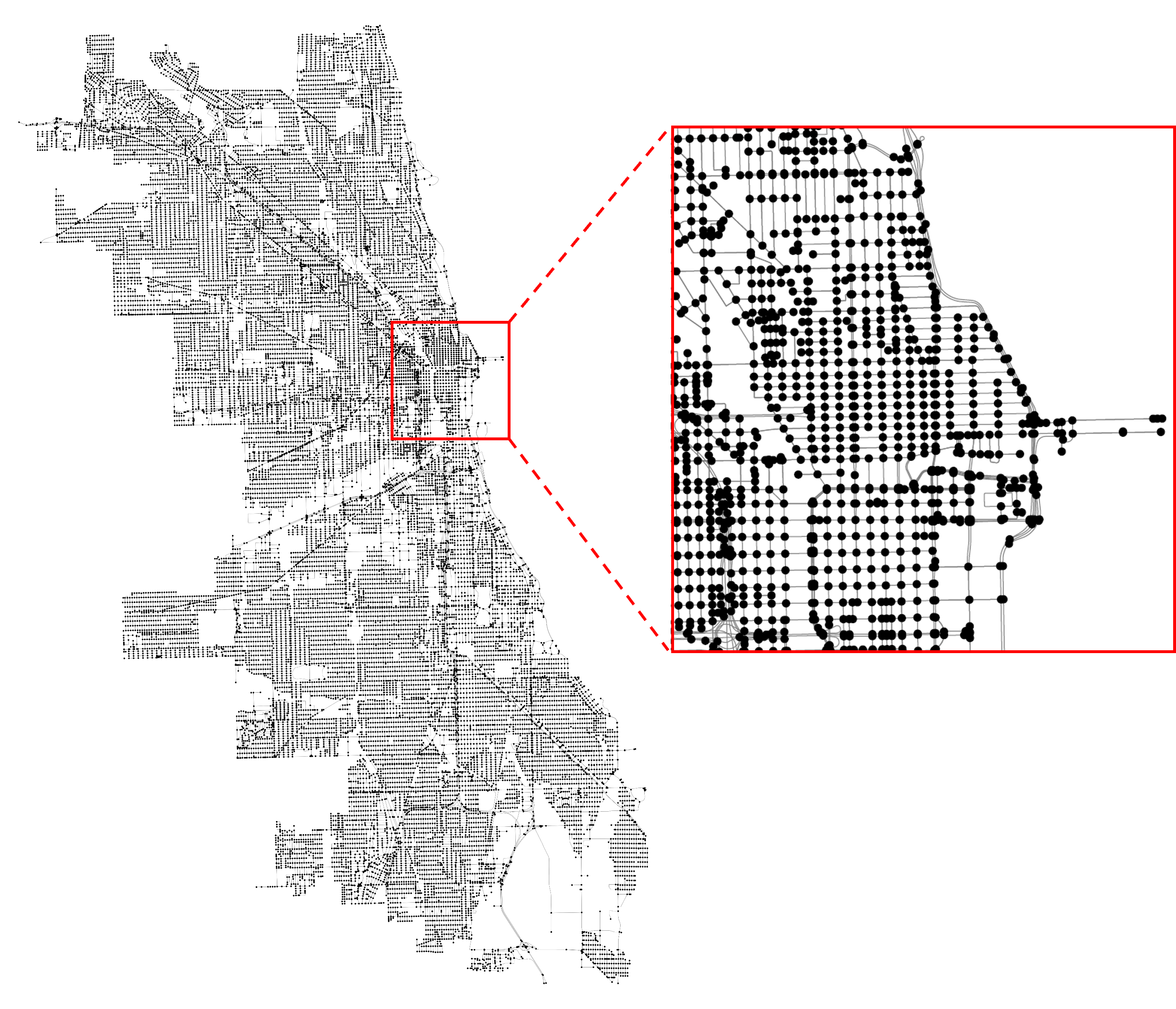}
    \caption{The simplified road network of Chicago using OSMnx. Nodes represent intersections and edges are road segments. The simplified road network shows the topology, but not the real road shape.}
    \label{fig:roadnet_chicago}
\end{figure}

We leveraged the utility of the OSMnx package to streamline and download road network data from the open-source platform, OpenStreetMap (OSM) \citep{boeing2017osmnx,haklay2008openstreetmap}. 
OSMnx operates by constructing a simplified and topologically corrected street network, effectively filtering out certain problematic elements such as edge dead-ends, edge self-loops, and complicated intersections where multiple streets intersect and at least one street continues beyond the intersection \citep{boeing2017osmnx,boeing2020multi,kirkley2018betweenness,ganin2017resilience,xue2022quantifying}. 
This process of simplification is important due to the inherent complexity of raw urban road network data sourced from OSM, which contains an overwhelming volume of information that often misrepresents the true topological relationship between intersections. 
For instance, OSM data employs multiple nodes and edges to represent a singular freeway, which can inadvertently compromise the performance of the GE model. 
Furthermore, additional complications arising from elements like dead-ends or intricately intertwined roads are also systematically eliminated through the process of simplification. 
Consequently, this transformative process significantly reduces the number of nodes (intersections) and edges (road segments) in the dataset. 
Following the implementation of OSMnx, the network was reduced from 390,642 nodes and 1,121,620 edges to 28,701 nodes and 76,174 edges. 
Figure \ref{fig:roadnet_chicago} visually depicts the simplified road network of Chicago after the implementation of OSMnx.

\subsubsection{Sociodemographic Data Description and Analysis}
\label{sec:socio_data}
The dataset central to this research is primarily derived from the American Community Survey (ACS) conducted over the span of 2017-2018. 
The ACS provides a wealth of sociodemographic information such as income brackets, age demographics, racial group compositions, prevailing modes of commuting, and average commuting durations, all specific to each of the 811 distinct census tracts located within the boundaries of Chicago. 

To further broaden the scope and depth of our analysis, we have also considered the potential benefits of manually engineered features derived from both road and transit networks, such as the number of rail stations and bus stops. 
These data have the potential to shed light on the capacity of Chicago's road network to manage varying volumes of traffic. 
Our inclusion of these road network features is based on precedents set by previous investigations that have successfully applied manual feature engineering to both the built environment and sociodemographic data in the context of mode share analysis \citep{wang2012understanding,zhan2017dynamics,li2015percolation,saberi2020simple}. 
For a comprehensive overview of the specific variables utilized in this research, refer to Table \ref{tab:chicago_variables}. 

We can categorize the sociodemographic inputs into the following classifications:
\begin{enumerate}
    \item Demographics (e.g., population, age, gender)
    \item Income and employment (e.g., income levels, employment status)
    \item Education (e.g., education levels, specific age and gender groups with certain educational attainments)
    \item  Race and ethnicity
    \item  Housing (e.g., property values, rent, housing units)
    \item  Area and density (e.g., area of the census tract, number of nodes per area)
    \item Transportation infrastructure (e.g., number of vehicles, vehicle per capita)
    \item Road Network features (e.g., road density, number of intersections)
\end{enumerate}

In our research, the travel mode shares under consideration encompass \textit{driving}, \textit{public transit (PT)}, \textit{taxi}, \textit{cycling}, and \textit{walking}. 
Shares of these modes can be derived from the travel\_ratio variables present in the sociodemographic data in Table \ref{tab:chicago_variables}, representing the choice set $C_n$. 
In Table \ref{tab:chicago_variables}, we detail a collection of 80 sociodemographic variables. 
Given the extensive list of variables, we employed a Pearson correlation coefficient with a threshold of 0.05 to refine our selection process. 
Only variables that demonstrated significant correlations ($\geq 0.05$) with all travel modes were retained to ensure their pertinence in our model. Variables that satisfy this threshold are colored in grey within the table.

\begin{table}[p]
    \scriptsize
    \centering
    \begin{tabularx}{\textwidth}{| X | X | X |}
    \toprule
        Census tract variables & Description & Variable level \\        
    \hline        
        \rowcolor[gray]{0.9} pop\_total & total number of population & Integer values $\in [347,20087]$ \\
        sex\_total & total number of population, same as pop\_total & Integer values $\in [347,20087]$ \\
        sex\_male & number of males & Integer values $\in [113,9179]$\\
        \rowcolor[gray]{0.9} sex\_female & number of females & Integer values $\in [201,11373]$\\
        age\_median & median age & Continuous values $\in [16,67.4]$\\
        households & number of households & Integer values $\in [113,12017]$\\
        race\_total & total number of population, same as pop\_total & Integer values $\in [347,20087]$ \\
        race\_white & number of white people & Integer values $\in [0,11933]$\\
        \rowcolor[gray]{0.9} race\_black & number of black people & Integer values $\in [0,6448]$\\
        race\_native & number of native American & Integer values $\in [0,294]$\\
        \rowcolor[gray]{0.9} race\_asian & number of asians & Integer values $\in [0,6166]$\\
        inc\_total\_pop & total number of population recoding incomes & Integer values $\in [263,17976]$ \\
        inc\_no\_pop & population without income & Integer values $\in [20,8718]$ \\
        inc\_with\_pop & population with income & Integer values $\in [180,17076]$\\
        inc\_pop\_10k & number of people with less than 10K income per year & Integer values $\in [21,1926]$\\
        \rowcolor[gray]{0.9} inc\_pop\_1k\_15k & number of people with more than 10K and less than 15K income per year & Integer values $\in [11,904]$\\
        \rowcolor[gray]{0.9} inc\_pop\_15k\_25k & number of people with more than 15K and less than 25K income per year & Integer values $\in [15,1412]$\\
        \rowcolor[gray]{0.9} inc\_pop\_25k\_35k & number of people with more than 25K and less than 35K income per year & Integer values $\in [5,980]$\\
        inc\_pop\_35k\_50k & number of people with more than 35K and less than 50K income per year & Integer values $\in [12,1377]$\\
        inc\_pop\_50k\_65k & number of people with more than 50K and less than 65K income per year & Integer values $\in [0,1830]$\\
        inc\_pop\_65k\_75k & number of people with more than 65K and less than 75K income per year & Integer values $\in [0,895]$\\
        inc\_pop\_75k & number of people with more than 75K income per year & Integer values $\in [0,8823]$\\
        \rowcolor[gray]{0.9} inc\_median\_ind & median individual income per year & Integer values $\in [4494,96667]$ \\
        \rowcolor[gray]{0.9} travel\_total\_to\_work & total number of population traveling to work & Integer values $\in [53,14332]$ \\
        travel\_driving\_to\_work & total number of population driving to work & Integer values $\in [32,6407]$\\
        travel\_pt\_to\_work & total number of population taking transit to work & Integer values $\in [8,4169]$ \\
        travel\_taxi\_to\_work & total number of population taking taxi to work & Integer values $\in [0,498]$\\
        travel\_cycle\_to\_work & total number of population riding bicycles to work & Integer values $\in [0,608]$ \\
        travel\_walk\_to\_work & total number of population walking to work & Integer values $\in [0,3756]$ \\
        edu\_total\_pop & total number of population (based on education) - not sure why it is different from pop\_total & Integer values $\in [200,17976]$ \\
        \rowcolor[gray]{0.9} bachelor\_male\_25\_34 & number of males with bachelor degree between 25 and 34 years old & Integer values $\in [0,1382]$ \\
        \rowcolor[gray]{0.9} master\_phd\_male\_25\_34 & number of males with master and PhD degree between 25 and 34 years old & Integer values $\in [0,1104]$ \\
        \rowcolor[gray]{0.9} bachelor\_male\_35\_44 & number of males with bachelor degree between 35 and 44 years old & Integer values $\in [0,443]$\\
        \rowcolor[gray]{0.9} master\_phd\_male\_35\_44 & number of males with master and PhD degree between 35 and 44 years old & Integer values $\in [0,977]$\\
        bachelor\_male\_45\_64 & number of males with bachelor degree between 45 and 64 years old & Integer values $\in [0,864]$ \\
        \rowcolor[gray]{0.9} master\_phd\_male\_45\_64 & number of males with master and PhD degree between 45 and 64 years old & Integer values $\in [0,1098]$\\
        bachelor\_male\_65\_over & number of males with bachelor degree older than 65 years old & Integer values $\in [0,286]$ \\
        master\_phd\_male\_65\_over & number of males with master and PhD degree older than 65 years old & Integer values $\in [0,566]$ \\
        \rowcolor[gray]{0.9} bachelor\_female\_25\_34 & number of females with bachelor degree between 25 and 34 years old & Integer values $\in [0,1175]$ \\
        \rowcolor[gray]{0.9} master\_phd\_female\_25\_34 & number of females with master and PhD degree between 25 and 34 years old & Integer values $\in [0,2007]$\\
        bachelor\_female\_35\_44 & number of females with bachelor degree between 35 and 44 years old & Integer values $\in [0,639]$ \\        
        \bottomrule
    \end{tabularx}
\end{table}

\begin{table}[htbp]
    \scriptsize
    \centering
    \begin{tabularx}{\textwidth}{| X | X | X |}
    \toprule
        Census tract variables & Description & Variable level \\        
    \hline
        \rowcolor[gray]{0.9} master\_phd\_female\_35\_44 & number of females with master and PhD degree between 35 and 44 years old & Integer values $\in [0,1154]$ \\
        bachelor\_female\_45\_64 & number of females with bachelor degree between 45 and 64 years old & Integer values $\in [0,788]$ \\
        master\_phd\_female\_45\_64 & number of females with master and PhD degree between 45 and 64 years old & Integer values $\in [0,996]$\\
        bachelor\_female\_65\_over & number of females with bachelor degree older than 65 years old & Integer values $\in [0,369]$\\
        master\_phd\_female\_65 & number of females with master and PhD degree older than 65 years old & Integer values $\in [0,575]$\\
        edu\_total & total number of population, a bit different from pop\_total & Integer values $\in [144,17171]$ \\
        \rowcolor[gray]{0.9} edu\_bachelor & number of people with bachelor degree & Integer values $\in [0,5322]$\\
        \rowcolor[gray]{0.9} edu\_master & number of people with master degree & Integer values $\in [0,4085]$\\
        \rowcolor[gray]{0.9} edu\_phd & number of people with PhD degree & Integer values $\in [0,1053]$\\
        inc\_median\_household & median household income & Integer values $\in [11146,194167]$\\
        \rowcolor[gray]{0.9} inc\_per\_capita & average income per capita & Integer values $\in [1801,134796]$\\
        employment\_total\_labor & total number of population (based on employment) & Integer values $\in [216,17976]$\\
        \rowcolor[gray]{0.9} employment\_employed & number of employed people & Integer values $\in [98,14680]$\\
        \rowcolor[gray]{0.9} employment\_unemployed & number of unemployed people & Integer values $\in [97,9362]$\\
        housing\_units\_total & total number of housing units & Integer values $\in [129,12660]$\\
        housing\_units\_occupied & total number of occupied housing units & Integer values $\in [113,12017]$ \\
        \rowcolor[gray]{0.9} housing\_units\_vacant & total number of vacant housing units & Integer values $\in [0,1580]$ \\
        \rowcolor[gray]{0.9} rent\_median & median ret & Integer values $\in [274,2563]$ \\
        \rowcolor[gray]{0.9} property\_value\_total & total property values & Integer values $\in [1,6612]$ \\
        \rowcolor[gray]{0.9} property\_value\_median & median property value & Integer values $\in [9999,1122700]$ \\
        \rowcolor[gray]{0.9} vehicle\_total\_imputed & total number of vehicles &  Integer values $\in [53,14332]$ \\
        household\_size\_avg & average household size & Continuous values $\in [1.3,35.6]$\\
        sex\_male\_ratio & ratio of males & Continuous values $\in [0,1]$ \\
        \rowcolor[gray]{0.9} race\_white\_ratio & ratio of white people & Continuous values $\in [0,1]$\\
        race\_black\_ratio & ratio of black people & Continuous values $\in [0,1]$\\
        race\_native\_ratio & ratio of native American & Continuous values $\in [0,1]$ \\
        race\_asian\_ratio & ratio of asians & Continuous values $\in [0,1]$ \\
        travel\_driving\_ratio & ratio of people driving to work & Continuous values $\in [0,1]$ \\
        travel\_pt\_ratio & ratio of people taking public transit to work & Continuous values $\in [0,1]$ \\
        travel\_taxi\_ratio & ratio of people taking taxi to work & Continuous values $\in [0,1]$ \\
        travel\_cycle\_ratio & ratio of people riding bicycles to work & Continuous values $\in [0,1]$ \\
        travel\_walk\_ratio & ratio of people walking to work & Continuous values $\in [0,1]$ \\
        travel\_work\_home\_ratio & ratio of people working from home & Continuous values $\in [0,1]$ \\
        \rowcolor[gray]{0.9}edu\_bachelor\_ratio & ratio of people with bachelor degree & Continuous values $\in [0,1]$ \\
        \rowcolor[gray]{0.9}edu\_master\_ratio & ratio of people with master degree & Continuous values $\in [0,1]$ \\
        \rowcolor[gray]{0.9} edu\_phd\_ratio & ratio of people with PhD degree & Continuous values $\in [0,1]$ \\
        \rowcolor[gray]{0.9} edu\_higher\_edu\_ratio & ratio of people with bachelor, master, or PhD degree & Continuous values $\in [0,1]$ \\
        \rowcolor[gray]{0.9} vehicle\_per\_capita & vehicle per capita & Continuous values $\in [0,1]$ \\
        vehicle\_per\_household & vehicle per household & Continuous values $\in [0,1]$ \\
        vacancy\_ratio & rato of vacant houses & Continuous values $\in [0,1]$ \\
        \rowcolor[gray]{0.9} area & Area of the census tract & Continuous values $\in [0.1, 38.8]$ (100sq km)\\
        num\_bus\_stop & \# of bus stops within each census tract & Integer values $\in [0,63]$ \\
        num\_rail\_stop & \# of rail stations within each census tract & Integer values $\in [0,18]$ \\
    \midrule
    \midrule
        Road network variables & Description & Variable level \\
    \hline
        road\_density & Road density defined by road length per area & Continuous values $\in [170.7, 18975.2]$ (km per 100sq km) \\
        \rowcolor[gray]{0.9} num\_node\_per\_area & \# of intersection per area  & Continuous values $\in [1.6, 97.3]$ (\# per 100sq km) \\
        num\_road\_per\_area & \# of road segments per area & Continuous values $\in [0.8, 189.5]$ (\# per 100sq km) \\
        sub\_sum\_cent & Summation of centrality for all intersections within each census tract & Continuous values $\in [1.0, 3.5]$ \\
        \rowcolor[gray]{0.9} sum\_deg & Summation of degrees for all intersections within each census tract & Integer values $\in [2, 858]$ \\
        sub\_sum\_nodes & Total \# of intersections within each census tract & Integer values $\in [2,166]$ \\
    \bottomrule
    \end{tabularx}
    \caption{Description of variables from the Chicago sociodemographic data and feature engineered road network variables.}
    \label{tab:chicago_variables}
\end{table}

We also include the features from network science and transit services. 
Variables such as $road\_density$ and $sub\_sum\_cent$ have been defined based on the frameworks proposed by \citet{hawbaker2005road} and \citet{van2013network} respectively. 
It is worth noting the presence of several features that exhibit substantial correlation with each other. 
For example, the calculation of $num\_node\_per\_area$ is based on $sub\_sum\_nodes$ and $area$. 
Moreover, racial group proportions are exclusive to each other, which underscores the necessity of exploring the degree of correlation between different features to accurately determine the inputs for the regression model. 
From Table \ref{tab:chicago_variables}, we can find that variables related to education and income exert more pronounced influences than other factors. 
Concurrently, younger age groups, specifically those aged 25-34 and 35-44, display a heightened impact on their travel mode decisions. 
Furthermore, distinct racial groups manifest varied influences on these choices, an issue that aligns with broader discussions on fairness and social equity within transit systems as elucidated by \citet{zheng2023fairness}.

\begin{figure}[htbp]
    \centering
    \includegraphics[width = 0.8 \textwidth]{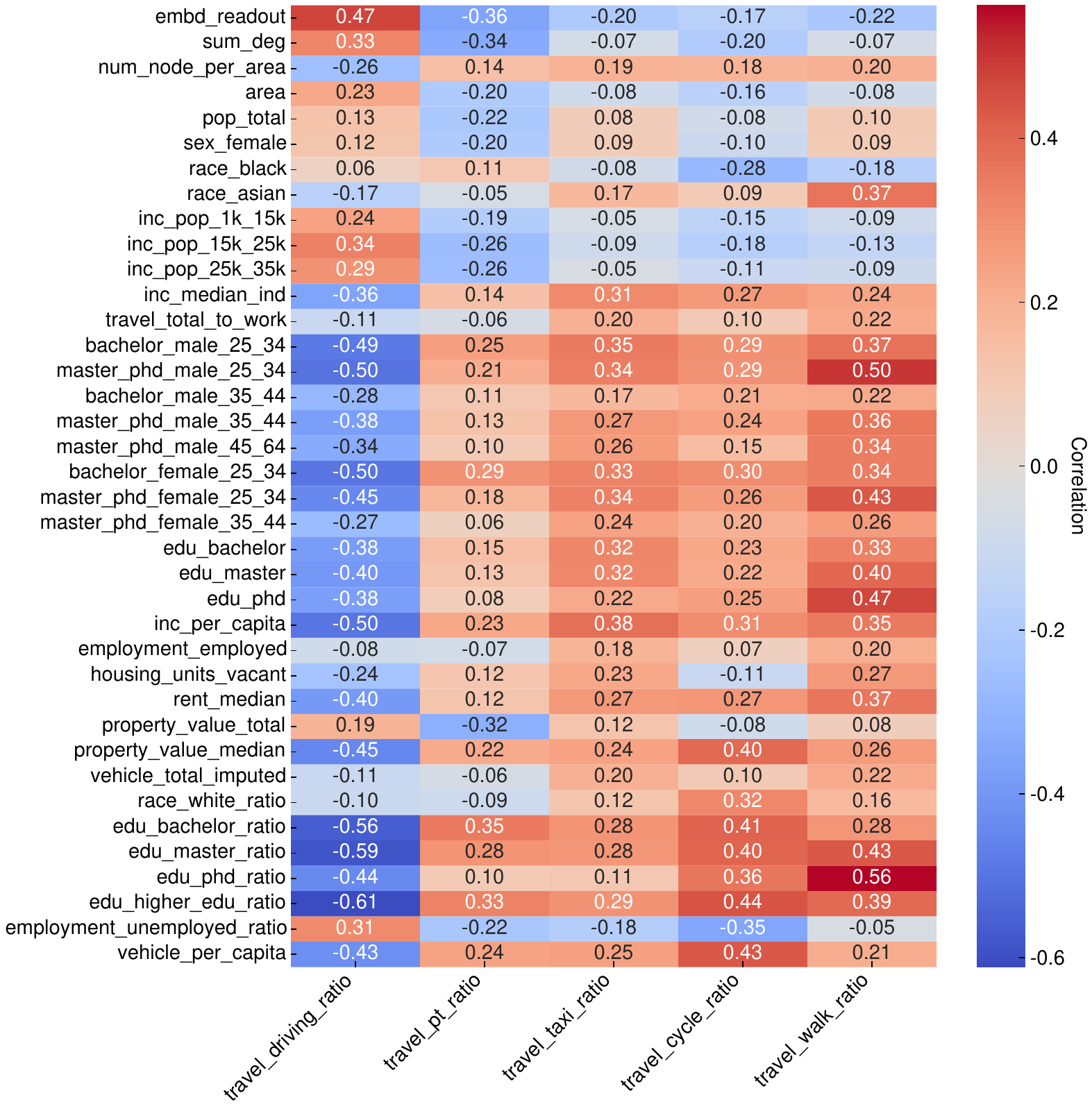}
    \caption{Correlation matrix of variables from the census survey, road network metrics, and graph embedding readouts.}
    \label{fig:corr}
\end{figure}

Figure \ref{fig:corr} shows the correlation matrix among variables. 
Notice that there is an additional variable denoted as \textit{embd\_readout}, which represents the average value of $\mathcal{R}_n$: $embd\_readout = \frac{1}{K} (\mathcal{R}_{n,1} + \mathcal{R}_{n,2} + \dots + \mathcal{R}_{n,K} )$. 
Given that all other variables constitute single values for each census tract, the computation of the average value is crucial for the correlation calculation. 
Meanwhile, number of rail stations and bus stops are also included to provide the insight from the transit service supply side.

Moreover, Figure \ref{fig:corr} reveals different levels of correlations in the input variables:
\begin{enumerate}
    \item Demographic variables tend to have a varied impact on travel mode shares. 
    Specifically, education level, especially for younger males and females, and race seem to play significant roles in influencing the choice of travel mode. 
    Younger and middle-aged males with a bachelor's or master's degree tend to drive less and prefer public transit, taxis, and cycling. 
    Areas with a higher black population have a slight correlation with a preference for walking and a negative correlation with cycling, and areas with a higher Asian population show a strong correlation with using taxis and walking.
    \item Income and employment variables significantly influence travel mode shares. 
    Lower income brackets (lower than 35k) tend to drive more and use public transit less. 
    Areas with higher median individual incomes tend to prefer public transit, taxis, cycling, walking, and working from home over driving. 
    Employed individuals show a preference for taxis and walking, whereas a higher unemployment ratio correlates with a preference for driving and a reduced preference for cycling. 
    \item Housing-related variables, especially concerning vacancy rates and rents, influence travel mode shares. 
    Areas with more vacant housing units tend to prefer taxis and walking over driving, while Areas with higher median rents have a negative correlation with driving and a positive correlation with public transit, taxis, cycling, and walking.
    \item  Higher education levels show a distinct preference for alternative travel modes to driving. Higher education ratios, including bachelor, master, and Ph.D., have a negative correlation with driving and a positive correlation with public transit, taxis, cycling, and walking. 
    \item The availability of vehicles in an area has a clear impact on driving mode share. It is interesting to see the \textit{vehicle\_per\_capita} has negative correlations with the driving modes. 
    \item The built environment of an area, including its size and infrastructure like road density and nodes, influence travel mode preferences. 
    Larger areas show a slight preference for driving and a negative correlation with public transit, taxis, cycling, and walking. 
    Areas with more intersections per area have a positive correlation with public transit, taxis, cycling, and walking. 
    Conversely, areas with higher road densities (more roads and intersections) tend to prefer driving and show a negative correlation with public transit and cycling.
    \item Our averaged graph embedding readout value, $embd\_readout$ has the strongest positive correlation with driving and the strongest negative correlation with public transit. 
    The exact nature and implications of this variable might be clearer with more context about what $embd\_readout$ represents in the dataset. 
    A critical clarification is that variables with lower correlations with the embedding readout values do not necessarily indicate the inadequacy of the readout for their estimation. 
    The variable $embd\_readout$ is essentially an aggregation of the GER, while the original GER vector with length $K$ encapsulates a greater amount of information useful in the regression.
\end{enumerate}

\subsection{Mode Share Prediction Performance}
Using the collected and refined feature data, our study aims to evaluate the numerical prediction capabilities of various models in relation to regressing the proportions associated with distinct travel modes. 
This assessment not only delves into the specifics of the inputs furnished to the model but also provides a comprehensive overview of the selected models for representation. 
Furthermore, to offer a holistic understanding, we will delineate the comparative outcomes derived from comparing the performance metrics of these models.

\subsubsection{Inputs and Travel Demand Models}
In order to evaluate the performance of DHMs, we measure the regression accuracy by employing different travel demand models, along with varying input parameters. 
Specifically, these inputs encompass the use of the 37 highlighted variables from Table \ref{tab:chicago_variables}, the sole employment of our urban road network GER, and a vector amalgamation of both. 
For ease of reference and clarity, we denote these three experimental conditions as the baseline, GER, and concatenated inputs, respectively. 

The reason why we need to include the concatenated inputs is that the objective of our DHM is to facilitate a comprehensive understanding of urban dynamics by concurrently utilizing both road network and sociodemographic data as inputs. 
As part of this study, we have previously expounded on the performance of GE as a standalone predictor. 
However, to enhance the model's predictive capability, we now propose a Concatenated input travel demand model, a model that concatenates these two distinct yet interrelated sources of data to enhance the prediction power.

We differentiate different input types.
We initiate our experimentation with the baseline inputs, representing a traditional mode share modeling methodology employed using sociodemographic and manually created network attributes as the inputs. 
In the context of our computational formulation, these values align with the $x_{ikn}$ parameters in Equation \ref{eq:origin}, embodying a linear utility structure. 
A granular breakdown of the 37-dimensional baseline input can be gleaned from Table \ref{tab:chicago_variables}, wherein the highlighted variables have been detailed discussed in Section \ref{sec:socio_data}. The graph embedding readouts, denoted as $\mathcal{R}_{n} \in \mathbb{R}^{1 \times K^*}, \forall n = 1,2, \dots, N$ is subsequently labeled as the GER input for models. The concatenation of the baseline and GER inputs is denoted as the concatenated input.

Note that our \textit{behavioral predictor}, denoted by $g(\cdot)$, boasts of inherent versatility, accommodating an array of prediction functions. 
Contextualizing this within our research spectrum, our predictive model $g$ integrates the following regression methodologies:

\textbf{MNL} is a sophisticated statistical tool tailored for analyzing choices among multiple discrete options, widely used in travel mode share analysis to determine the probability of selecting various transportation options like car, bus, or bike. 
Distinguished by its capacity to handle multiple categorical outcomes, it employs a range of predictors—socioeconomic profiles, trip details, and transportation system features—to compute the probabilities of each mode being chosen. 
This model yields insights into how changes in predictors affect the odds of selecting each travel mode, thereby aiding in predicting mode share and guiding transportation policy decisions \citep{hausman1984specification,anas1981estimation}.

\textbf{Random Forests}, transitioning from traditional regression to ensemble methods, it employ multiple decision trees to yield predictions with heightened accuracy and resilience \citep{biau2016random}. 
In the context of travel mode share analysis, their potency is particularly evident. 
They capably manage extensive predictor sets, account for intricate variable interactions, and mitigate potential overfitting inherent to a singular decision tree. 
When forecasting mode shares, the decision trees collectively vote for a travel mode based on input features, with the mode securing the majority being the conclusive prediction. 
The breadth of these features mirrors those in logistic regression, encompassing socio-economic traits and transportation specifics.
A salient advantage of random forests is their capacity to prioritize predictors, guiding researchers and planners to discern paramount factors steering travel mode decisions. T
his knowledge becomes foundational in sculpting potent transportation policies and interventions.

\textbf{XGBoost}, standing for "Extreme Gradient Boosting", represents the cutting edge in gradient boosting frameworks, distinguished by its prowess and efficiency in both competitive and practical machine learning scenarios \citep{,chen2015xgboost}. 
Rooted in gradient boosting principles, XGBoost creates a cumulative ensemble of decision trees, with succeeding trees remedying the predecessors' errors. 
In the sphere of travel mode selection and mode share prediction, XGBoost excels by capturing intricate non-linear associations and variable interactions, often outclassing many conventional algorithms in predictive accuracy. 
Key attributes such as adeptness at navigating missing data, intrinsic regularization to counteract overfitting, and scalability make it a frontrunner for extensive transportation data analytics. 
Complementing its predictive capabilities, XGBoost's feature importance metrics furnish researchers and planners with invaluable insights into pivotal determinants shaping travel mode inclinations, thereby informing strategic transport policy crafting and infrastructure development endeavors.

We fine-tune both the Random Forest and XGBoost model parameters based on their performance in regressing each of the travel modes. 
That means we try multiple combinations of parameter sets and always present the best model performances when regressing the travel mode share.

\subsubsection{Model Comparison}
Before introducing the comparison results, we firstly specify our data split approach. 
We then describe the chosen evaluation metrics and present the numerical analysis of mode share regression results.

We split the 811 census tracts of Chicago by random 70-30 train-test split. 
The different input formats are divided accordingly. 
The metrics we use are in-sample R-square (ISR2) and out-of-sample R-square (OSR2), which are the linear regression models run on the train and test set respectively. 

The R-square value, often termed the coefficient of determination, quantifies the proportion of the variance in the dependent variable that is predictable from the independent variables. It is defined as:
\begin{equation}
    R^2 = 1 - \frac{\sum_i (y_i - f_i)^2 }{\sum_i (y_i - \frac{1}{N} \sum_{n=1}^N y_i )^2 }
\end{equation}

\noindent where $y_i$ and $f_i$ are the true values and regression results accordingly. 
The ISR2 value provides insight into the model's performance on the data it was trained on. 
A higher ISR2 value suggests that the model can explain a significant portion of the variability in the dependent variable based on the training dataset. 
However, an excessively high ISR2 can be indicative of potential overfitting, suggesting the model might be excessively tailored to the training data and could underperform when exposed to new, unseen data. 
On the other hand, the OSR2 value evaluates the model's predictive capacity on previously unseen data, in this context, the test set. 
This metric is pivotal as it assesses the model's ability to generalize to new data. 
A pronounced disparity between ISR2 and OSR2 values can signal overfitting or underfitting. This underlines the significance of assessing models on both training and testing datasets to guarantee their robustness and predictive precision.

Table \ref{tab:model_comparison_new} provides a comprehensive comparison of three different modeling techniques applied to travel mode share analysis using three different inputs:

\begin{table}[htbp]
    \footnotesize
    \centering
    \begin{tabularx}{\textwidth}{cX|*{6}{>{\centering\arraybackslash}X}}
    \hline
       \multirow{2}{*}{Models \& Modes} &  & \multicolumn{2}{c}{Baseline input} & \multicolumn{2}{c}{GER input} & \multicolumn{2}{c}{Concatenated input}\\
       \cline{3-4}\cline{5-6}\cline{7-8}
       & & ISR2 & OSR2 & ISR2 & OSR2 & ISR2 & OSR2 \\
       \hline
       \multirow{6}{*}{MNL}  & Driving &  0.752 & 0.679  &  0.814 & \underline{0.708} &  0.858 & \textbf{0.747 } \\
       & PT &  0.546 & 0.388  &  0.731 & \underline{0.553}  &  0.785 & \textbf{0.588}  \\
       & Taxi &  0.241 & 0.203  & 0.441 & \textbf{0.366} &  0.503 & \underline{0.330}  \\
       & Cycling &  0.368 & 0.231  &  0.616 & \underline{0.279}  &  0.656 &  \textbf{0.390} \\
       & Walking &  0.642 & 0.512  &  0.731 & \underline{0.624}  &  0.855 & \textbf{0.646}  \\
    \hline
       \multirow{6}{*}{Random Forest} & Driving  &  0.939 & 0.559  & 0.959 & \textbf{0.697} &  0.955 & \underline{0.686}  \\
       & PT  &  0.904 & 0.470  &  0.931 & \underline{0.562}  &  0.938 & \textbf{0.574}  \\
       & Taxi  &  0.515 & 0.166   &  0.684 & \textbf{0.292}  &  0.587 & \underline{0.242} \\
       & Cycling  &  0.837 & 0.380  &  0.877 & \underline{0.389}  &  0.882 & \textbf{0.466}  \\
       & Walking  &  0.873 & 0.344  &  0.915 & \underline{0.496}   &  0.937 & \textbf{0.519}  \\
    \hline
        \multirow{6}{*}{XGBoost} & Driving &  0.961 &   0.622  &  0.995 & \underline{0.727}&  0.997 & \textbf{0.753}  \\
        & PT &  0.935 & 0.549  &  0.972 & \underline{0.576}  & 0.996  & \textbf{0.639}  \\
        & Taxi &  0.534 & 0.200  &  0.693 & \underline{0.318} &  0.736 & \textbf{0.35}  \\
        & Cycling &  0.852 & 0.144  &  0.964 & \underline{0.168} &  0.993 & \textbf{0.332}  \\
        & Walking &  0.975 & 0.435  &  0.986 & \underline{0.533}  &  0.991 & \textbf{0.553}  \\
    \hline
    \end{tabularx}
    \caption{Travel demand model comparisons on different travel modes using different input features. Bold fonts indicate the best out-of-sample regression results for a specific travel mode across all the travel demand models. The underlining indicates the best ISR2 or OSR2 for a particular travel model using a particular model.}
    \label{tab:model_comparison_new}
\end{table}

From the analysis, it's evident that models utilizing concatenated inputs consistently outshine others, particularly in predictions pertaining to unseen data. 
This underscores the value of blending conventional data with insights from urban layouts to more accurately anticipate travel behaviors. 
On average, the introduction of GER inputs yields a 20\% enhancement in OSR2 relative to baseline models. 
Furthermore, the top-performing results derived from either the GER-input travel demand models or the Concatenated travel demand model demonstrate a remarkable 40\% improvement in OSR2 when contrasted with the baseline models.

When we look at individual travel modes:
\begin{itemize}
    \item For Driving, models using concatenated inputs consistently outperformed the others, suggesting that understanding both personal factors and city layout is crucial for predicting driving behavior.
    \item For PT and Walking, there was a clear benefit from using GER and concatenated inputs, underlining the importance of the urban environment in influencing these choices.
    \item Taxi and Cycling predictions also saw improvements with GER and concatenated inputs, indicating that the city's road structure can impact decisions to hail a taxi or hop on a bike.
\end{itemize}

As for the predictive models, MNL didn't predict as accurately as the ensemble models, particularly when enriched with GER and concatenated inputs. 
Ensemble models like Random Forest and XGBoost, especially when input with the concatenated dataset, achieve higher OSR2. 
However, a key observation was the occasional gap between a model's predictions for its training data and its predictions for new data. 
This was especially noticeable for the Random Forest model using just the baseline inputs. 
This might mean the model is too tailored to its training data, which could make it less accurate in real-world scenarios.

\begin{figure}[hp]
    \centering
    \subfigure[PT Mode Share]{\includegraphics[width=0.9\textwidth]{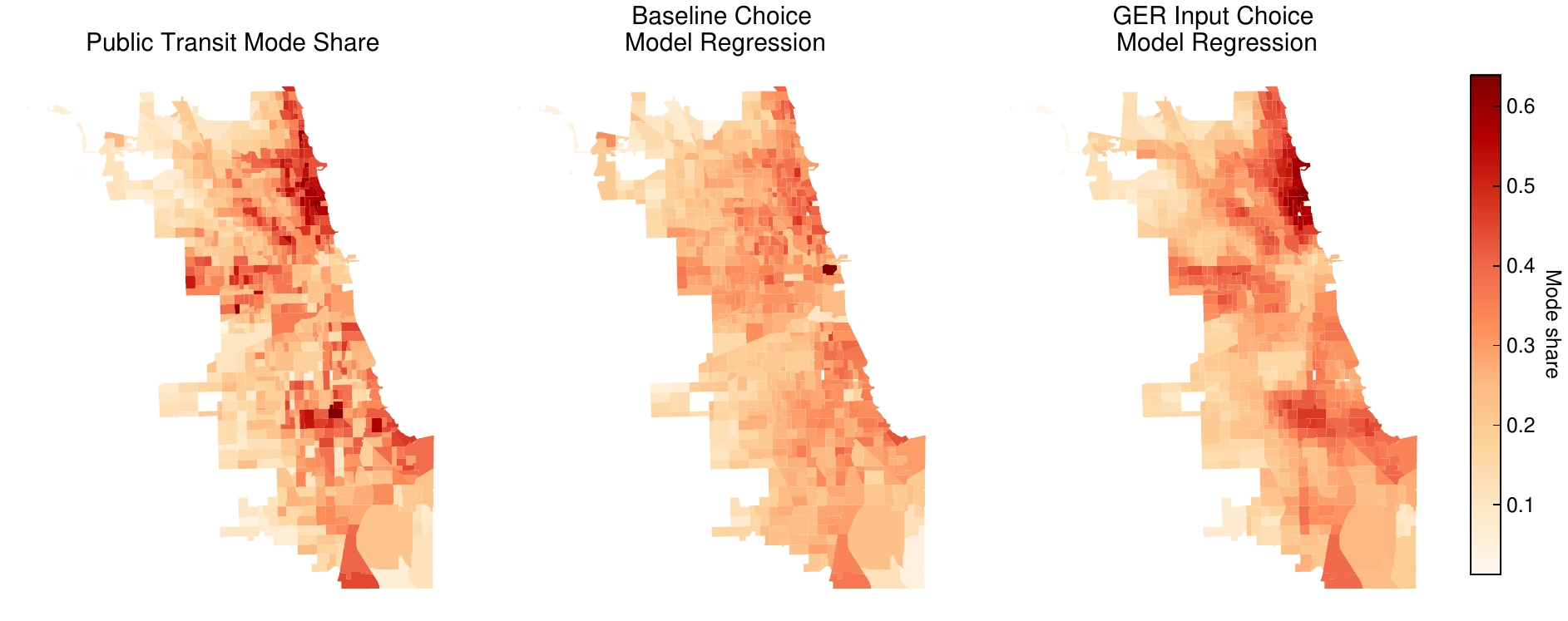}
    \label{fig:viz_pt}}
    
    \subfigure[Driving Mode Share]{\includegraphics[width=0.9\textwidth]{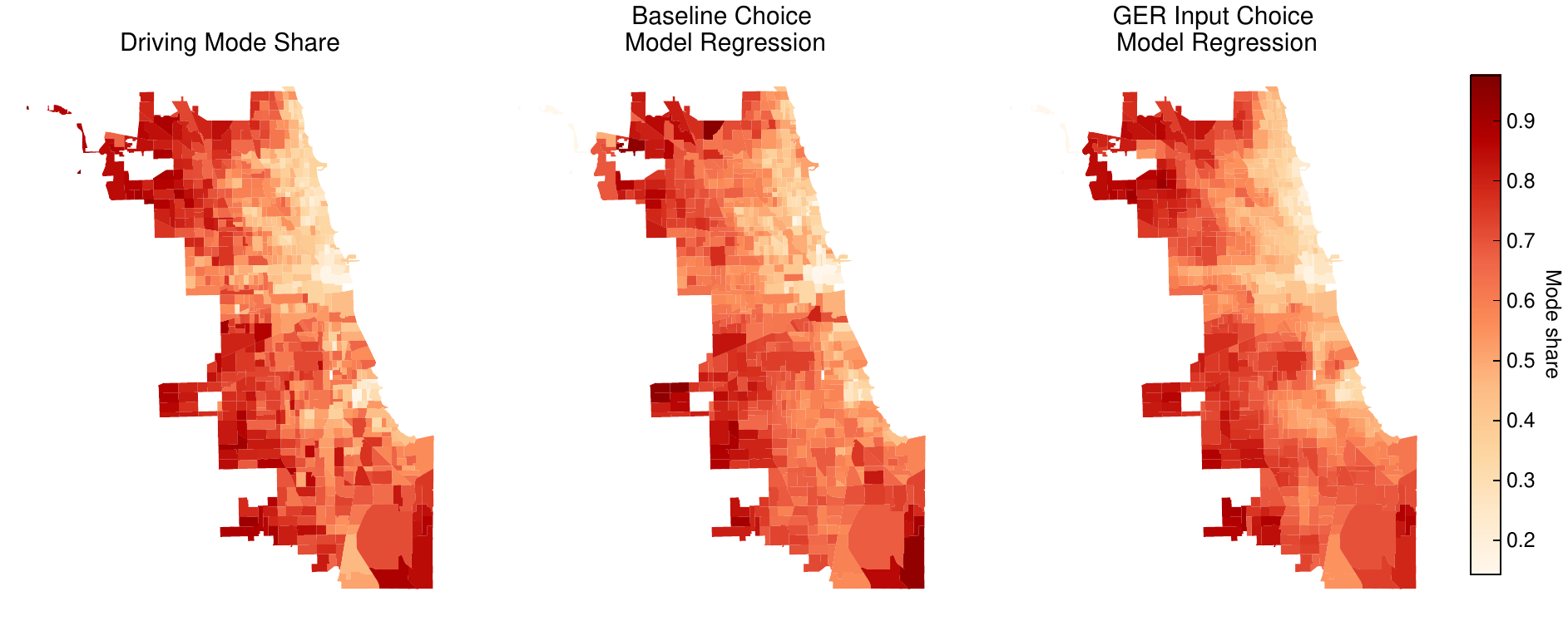}
    \label{fig:viz_driving}}

    \subfigure[Taxi Mode Share]{\includegraphics[width=0.9\textwidth]{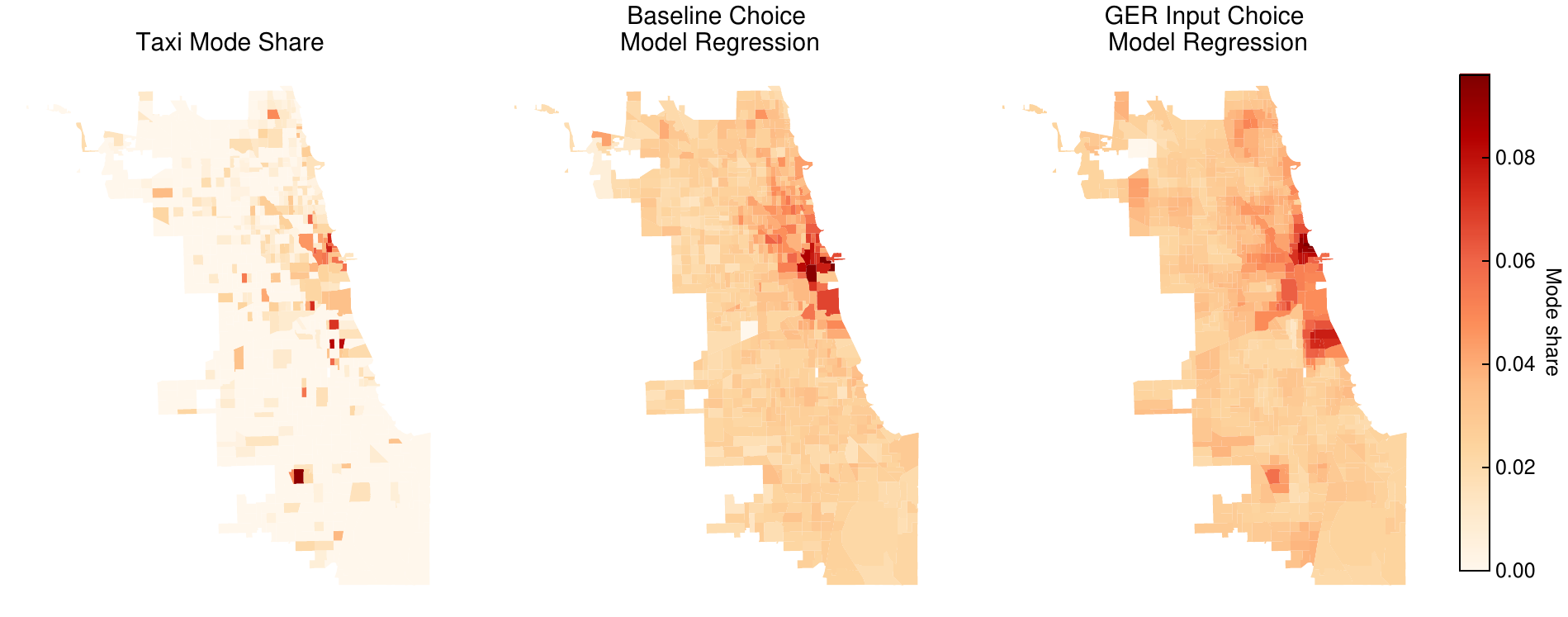}
    \label{fig:viz_taxi}}

    \caption{Comparison of different mode share regression outcomes (part 1): Ground truth vs. baseline and GER-input demand models using MNL}
    \label{fig:viz}
\end{figure}

\begin{figure}[hp]
    \centering
    \subfigure[Cycling Mode Share]{\includegraphics[width=0.9\textwidth]{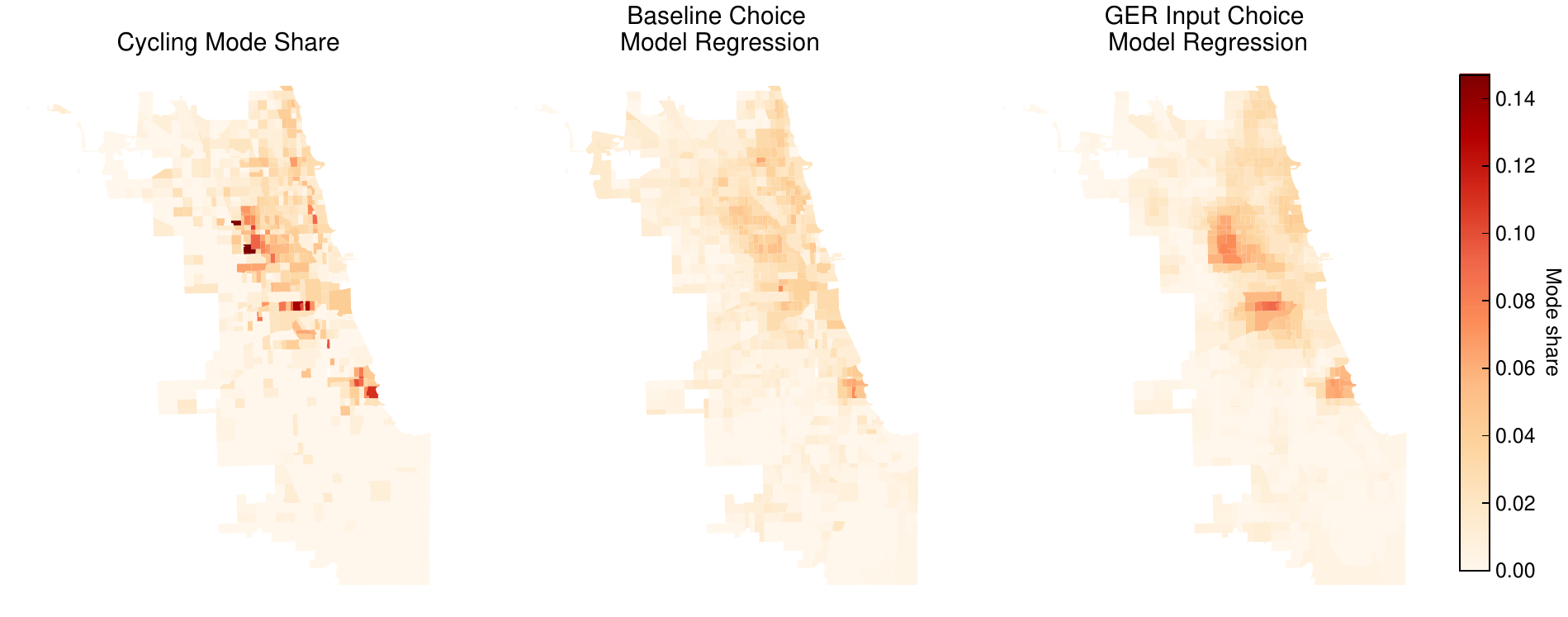}
    \label{fig:viz_cycle}}

    \subfigure[Walking Mode Share]{\includegraphics[width=0.9\textwidth]{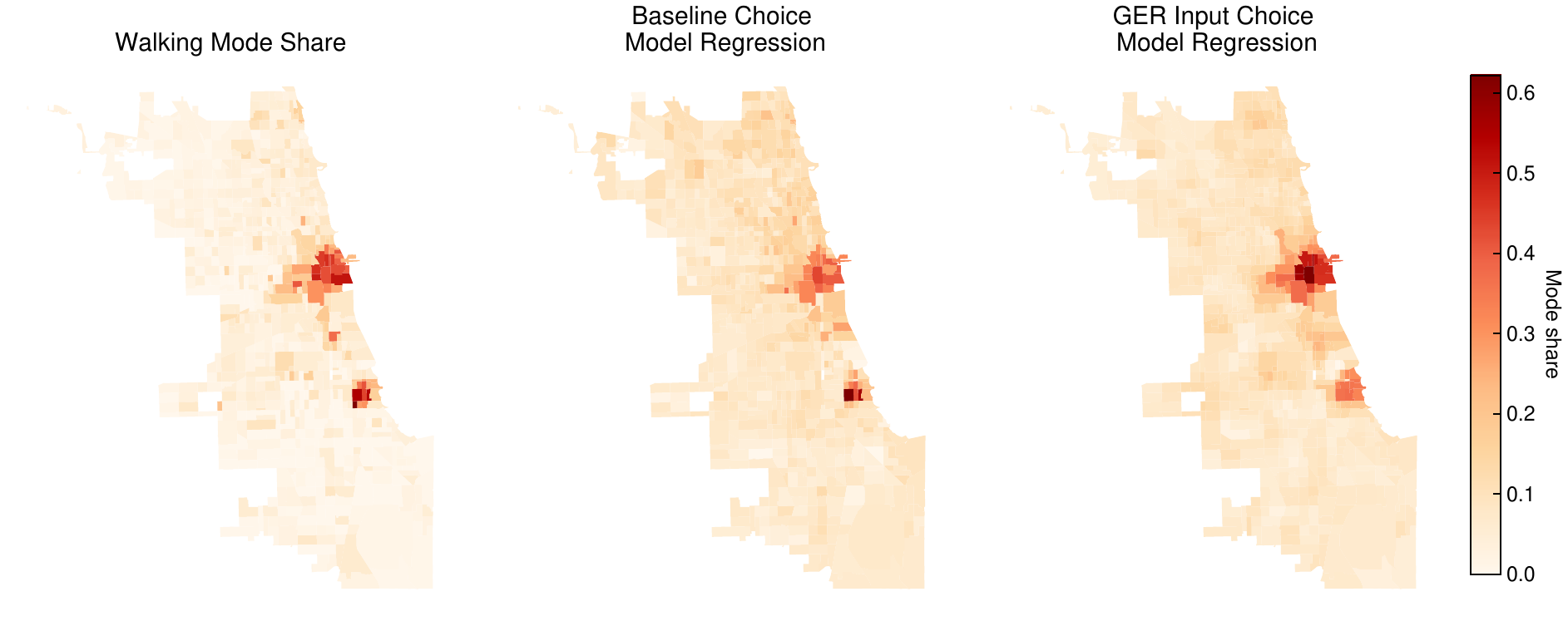}
    \label{fig:viz_walk}}

    \caption{Comparison of different mode share regression outcomes (part 2): Ground truth vs. baseline and GER-input travel demand models using MNL.}
    \label{fig:viz_part2}
\end{figure}


\section{Visual Interpretation}
\label{sec:interpretation}
\subsection{Spatial Patterns of Regression Results}
The findings presented in Table \ref{tab:model_comparison_new} shed light on a rather interesting finding. 
It is evident that when the road network topology is employed solely as the defining input, the GER-input demand model substantially outstrips the performance of the Baseline demand model, registering a commendable improvement margin of approximately 20\%. 
This pivotal outcome underscores the importance of delving deeper into the inherent patterns and discernible physical implications that the embeddings might elucidate. 
In pursuit of a comprehensive comparative analysis of the spatial distributions associated with regression outcomes across the entire city, we opted to re-conduct the fitted linear regression model, this time leveraging the entirety of the available data.

The regression results for each of the travel modes are visualized spatially in Figure \ref{fig:viz} and \ref{fig:viz_part2}. 
We present the results derived from MNL, as predicted either by the Baseline Demand Models or the GER-input Demand Models.

In examining the spatial distributions of various transportation mode shares across Chicago, several distinct patterns emerge when comparing the ground-truth data with predictions from the baseline and GER models. 
From Figure \ref{fig:viz_pt}, it is clear that public transit mode shares exhibit a lack of spatial contiguity in relation to their neighboring regions in numerous locales. 
While the Baseline demand model predominantly tends towards underfitting the data, it remarkably retains the capability to assimilate the overarching spatial contour. 
Contrarily, our GER-input demand model augments the linear model's capacity to differentiate nuances and, in doing so, unveils a pronounced spatial continuity.

Conversely, Driving mode share patterns in Figure \ref{fig:viz_driving}, both models exhibit broadly congruent patterns, capturing the prevalent driving behaviors of Chicago's inhabitants. 
However, subtle variations, particularly in the intensity and localization of driving patterns, accentuate the GER model's heightened sensitivity. 
For Cycling mode share pattern in Figure \ref{fig:viz_cycle}, the GER model aligns superiorly with the ground-truth data, capturing urban transit dynamics effectively. 
Specifically, the GER's spatial representation closely matches the actual distribution, while the baseline model appears less connected to these dynamics. 
The Taxi and Walking Mode Shares present intriguing contrasts, shown in Figure \ref{fig:viz_taxi} and \ref{fig:viz_walk}. 
For taxi usage, the GER model portrays a more dispersed and pronounced distribution, suggesting its potential to discern intricate transit choices. 
In terms of walking, while both models convey analogous distributions, the GER model presents a more resonant depiction, especially in pedestrian-dominant regions.

This naturally leaves us with the question: how exactly does the GER model's performance interface with sociodemographic data and the intricate constructs of road networks? 
We therefore further interpret the patterns of $\mathcal{R}$ and its correlations with other features.

\subsection{Relation Between Graph Embedding Readouts and Road Network Structures}

\begin{figure}[t]
    \centering
    \subfigure[Value of $embd\_readout$]{\includegraphics[width=0.3\textwidth]{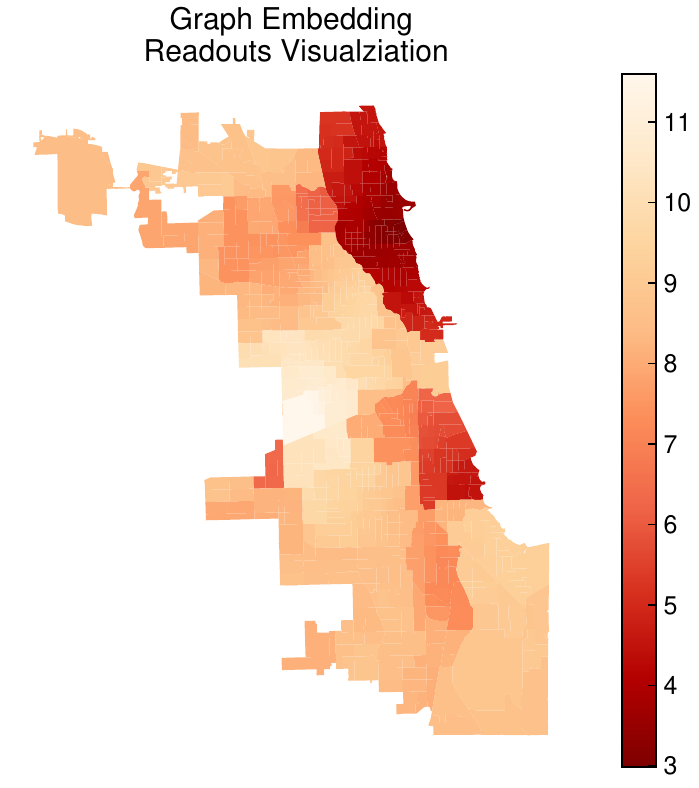}}\hspace{5em}
    \subfigure[Chicago boroughs]{\includegraphics[width=0.3\textwidth]{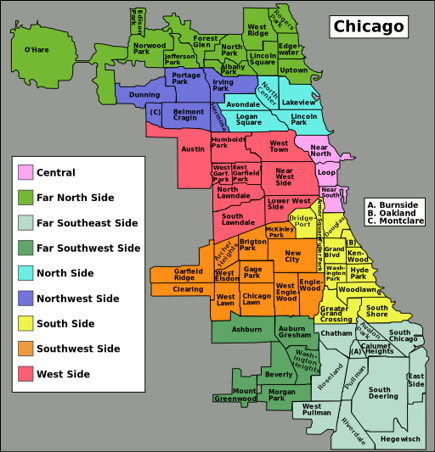}}
    \caption{Visualization of averaged GER values along and the boroughs of Chicago. The North Side, South Side, and Central parts of Chicago are typically considered popular areas.}
    \label{fig:pattern_chicago}
\end{figure}

In an effort to visualize the spatial implications and insights unveiled by the graph embedding readouts, we employ the value of $embd\_readout$ to investigate the correlations with the geographical characteristics of the city. 
The geographical distribution of $embd\_readout$ demonstrates spatial continuity, depicted in Figure \ref{fig:pattern_chicago}. 
It is intriguing to observe that elevated values of $embd\_readout$ are predominantly found in the Northern and Southern regions of Chicago, areas recognized for their prosperity and popularity within the city's confines \footnote{\url{https://en.wikipedia.org/wiki/Community_areas_in_Chicago}}. 
This observation intriguingly aligns with the robust negative correlation between $embd\_readout$ and the public transit mode share, as elucidated in Figure \ref{fig:corr}.

As we delve deeper into the interpretation of the graph embedding readouts, we aspire to visualize the structure of the road network within each census tract in correspondence with the sorted graph embedding readouts. 
Given the 811 census tracts, each is assigned a graph embedding value to facilitate the visualization as in Figure \ref{fig:pattern_chicago}. 
This enables us to sort these values, facilitating the identification of the road network that corresponds to the ${5\%, 25\%, 50\%, 75\%, 95\%}$ quantiles of the graph embedding readouts. 
By doing this, we aim to elucidate the correlations between the numerical graph embedding readouts and the inherent network topology of the inputs. 
The result of this exploration is illustrated in Figure \ref{fig:progressive}:

\begin{figure}[H]
    \centering
    \includegraphics[width = 0.7\textwidth]{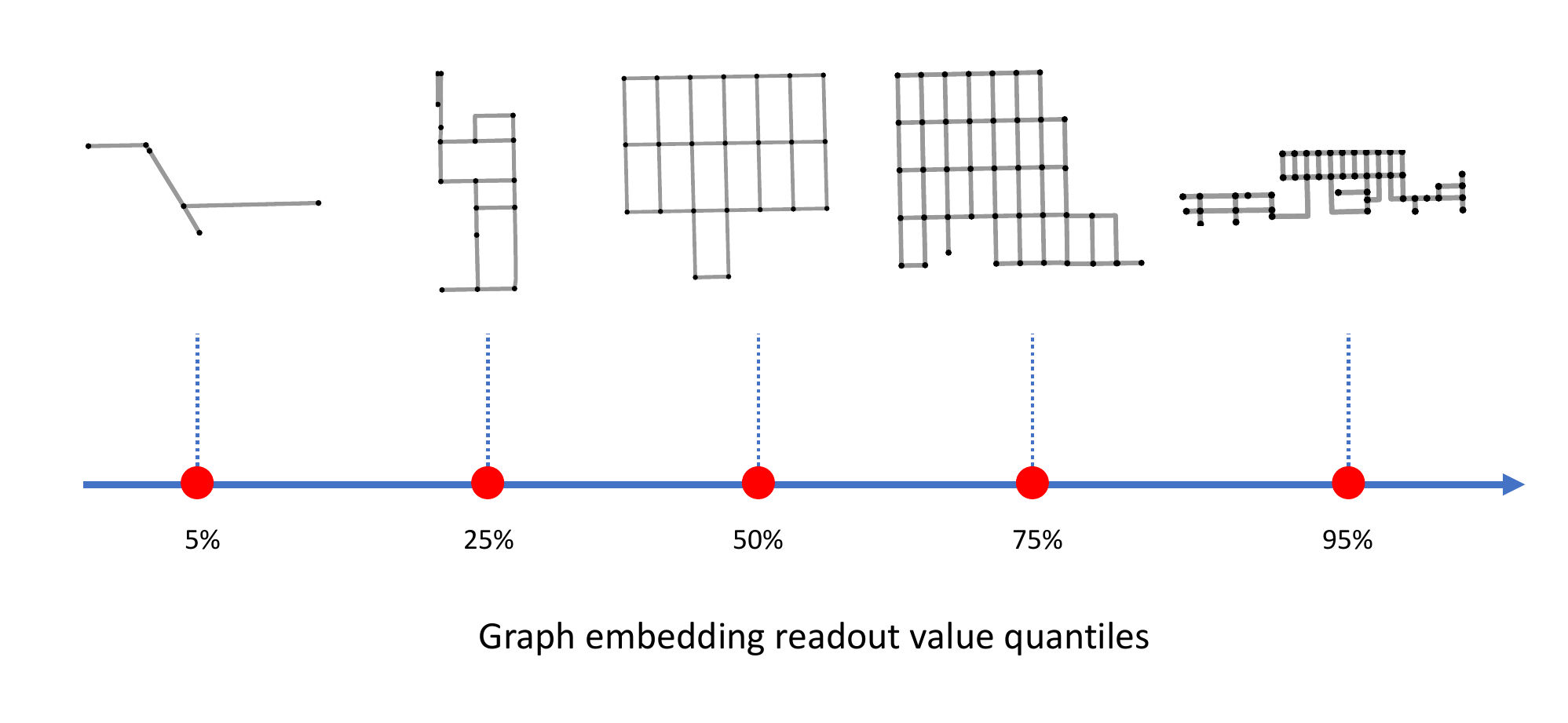}
    \caption{Quantiles of graph embedding readouts and their corresponding road network structures.}
    \label{fig:progressive}
\end{figure}

Figure \ref{fig:progressive} presents an evolution from smaller to larger quantiles, wherein the depicted road network structure transitions from a sparse, irregular, and non-rectangular form to a denser, organized, and grid-like structure. 
This evolution is concordant with the underlying logic of the GE technique. 
The technique relies on formulating the information passing in the network, necessitating more weight/values on popular nodes to encapsulate the complexities of network structures. 

In the context of real-world applications, denser and more organized road structures typically signify areas of high travel demand, such as downtown districts or central business sectors. 
Thus, along with the insights gleaned from Figure \ref{fig:pattern_chicago}, it becomes apparent that GER have substantial potential in estimating aspects such as the density, popularity, prosperity, and other sociodemographic features of a city.

\subsection{Clustering Analysis of the GER}

The GE technique possesses a potent capability to effectively discern the relationships between sociodemographic factors and census tracts, utilizing only road network topology as its primary input. 
This process is notable for its capacity to distill the heterogeneous information derived from different sociodemographics into a more coherent and concise representation of census tracts. 
Consequently, it unveils intricate correlations among these tracts, obviating the need for laborious collection and analysis of various sociodemographic factors.

To further elucidate the spatial correlation patterns of these embeddings, we employ clustering algorithms to visually demarcate areas of similar characteristics. 
Specifically, for this study, we have utilized the K-Means Clustering Algorithm \citep{krishna1999genetic} on the graph embedding readouts. 
The decision to opt for 30 clusters was dictated by the fact that it constitutes roughly 30\% of the total number of census tracts, thereby ensuring a sufficient level of granularity in the spatial division. 
Of course, it's worth mentioning that other clustering algorithms, as well as different numbers of clusters, could also be employed depending on the specific analytical requirements and data characteristics.

\begin{figure}[htbp]
    \centering
    \includegraphics[width = 0.4\textwidth]{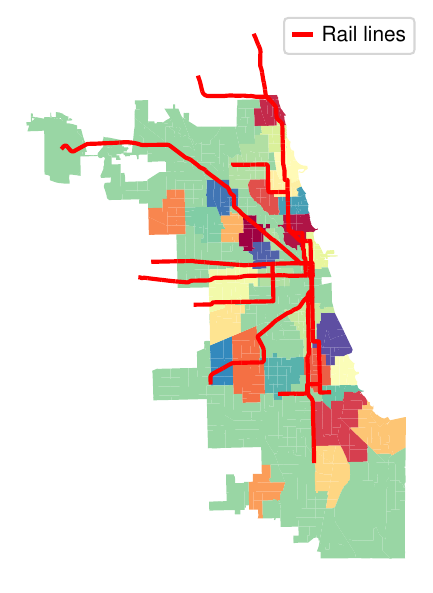}
    \caption{30 clusters generated by applying Gaussian Mixture Models Clustering Algorithm on GER. Red lines represent the distribution of rail lines.}
    \label{fig:cluster}
\end{figure}

The spatial distribution of the resulting clusters is depicted in Figure \ref{fig:cluster}. 
Each unique color in Figure \ref{fig:cluster} symbolizes a distinct cluster, with the largest cluster manifesting itself in light green. 
Upon examining these clusters, it's apparent that they closely reflect the geographical proximity of neighboring census tracts. 
The GERs possess the capacity to learn not only the intricacies of network structures but also their spatial distances, thereby demonstrating a comprehensive understanding of the spatial relationships within the data. 
Interestingly, the clusters are generally observed to align with the city's rail lines, implying a strong correlation between public transit infrastructure and the spatial distribution of these tracts.

Such a finding significantly emphasizes the profound impact that the transit system can exert on various aspects of urban life. 
Not only does it shape the physical structure of the road network, but it also influences a myriad of social characteristics within its sphere of influence. 
These may encompass income levels, age demographics, and transit usage patterns among the resident population. 
Thus, this observation offers valuable insights into the interplay between infrastructure planning and sociodemographic development, further attesting to the applicability and efficacy of our approach in urban studies.

\subsection{Interpretation of the GER variables}

Understanding the relative importance of distinct attributes is important in mode share analysis. 
Typically, elasticity analysis, which measures the effect of a 1\% change in an independent variable on a dependent variable, has been the standard in traditional choice models. 
This relies on evaluating and interpreting each input dimension. 
Within the ambit of DHMs, however, the normalization which bounds the embedded values to the [0,1] range intimates that changes in embedded values can't be directly attributed to fluctuations in a singular dimension. 
The crux of this limitation lies in the embedded values’ representation: they are but a specific output from the GE model and lack the tangible significance synonymous with input values in baseline demand models. 
Thus, a mere 1\% perturbation in embedding values might be devoid of substantive interpretation.

\begin{figure}[htbp]
    \centering
    \includegraphics[width = 0.8\textwidth]{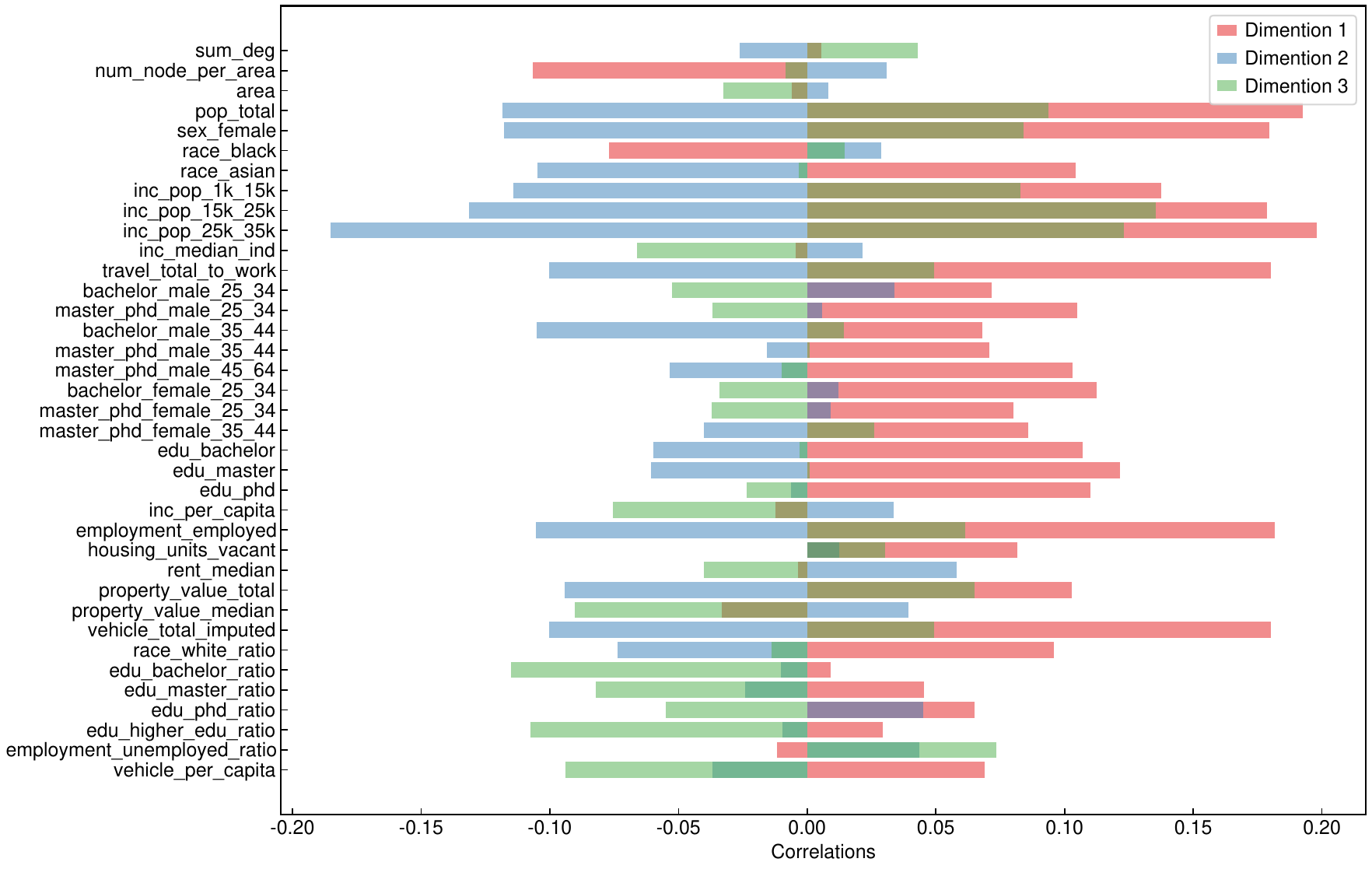}
    \caption{Interpretations of latent variables.}
    \label{fig:interpretation}
\end{figure}

Yet, understanding the relative weight of each dimension is achievable. 
We investigate the correlations between each embedded dimension and assorted sociodemographic indicators. 
Referring to the provided Figure \ref{fig:interpretation}, the correlations between the initial three dimensions of the embedded values and various sociodemographic factors are highlighted. 
These correlations provide us with the analytical perspectives of how each dimension might resonate with the influences of the embedded value dimension.

From the visual data shown in Figure \ref{fig:interpretation}, Dimension 1 appears to have a negative association with attributes such as \textit{sum\_deg} and \textit{num\_node\_per\_area}, whereas it establishes a positive correlation with indicators like \textit{pop\_total} and \textit{sex\_female}. 
Such correlations suggest that this dimension may primarily be inclined towards understanding population metrics and gender distribution. 
In contrast, Dimension 2 presents strong correlations with factors like \textit{race\_black}, \textit{race\_asian}, and \textit{inc\_pop\_1k\_15k}, signifying its sensitivity towards racial demographics and certain income brackets. 
Lastly, Dimension 3, while exhibiting less pronounced associations with racial metrics, underscores its relevance with economic parameters like \textit{inc\_median\_ind} and \textit{travel\_total\_to\_work}, pointing towards an economic or occupational-oriented theme.

While this interpretation approach echoes the methodology of topic modeling, it is invariably constrained by the scope of human prior knowledge and understanding. 
Future efforts to enhance interpretation could consider incorporating generative models to reverse the GE process into graph generation. 
This approach would utilize the GER values as inputs and the road networks as outputs, which would foster a more nuanced understanding of how each dimension influences the learning of road network structures.
 
\section{Conclusion}
\label{sec:conclusion}
In this paper, we have developed a novel framework by integrating graph-embedded urban road networks to train sophisticated travel demand models for mode share analysis, obviating the need for exhaustive feature engineering or reliance on a priori knowledge of built environment. 
This approach entails the introduction of the DHM, a computational framework that fuses the principles of deep learning and hybrid choice models, underpinned by graph embedding techniques. 
Specifically, the GE method, \textit{Node2Vec}, is utilized to predict the individual portions of choosing different travel modes. 
Our empirical results substantiate the assertion that the implementation of GE can considerably amplify the effectiveness of mode share analysis. 
Furthermore, the amalgamation of GERs with sociodemographic variables demonstrably enhances the overall performance of the regression model. 
The DHM framework is adaptive to other machine learning predictive methods. 
We further recognize that the GE technique offers profound spatial insights, which are inherently correlated with the popularity and affluence of various census tracts. 
This correlation further extends to the density and morphological characteristics of road networks within respective regions. 
Furthermore, the GER clusters evince robust spatial contiguity and a notable convergence around railway lines.

Looking ahead to future research directions, we aspire to bolster the interpretability and generalizability of the DHM. 
A promising direction involves incorporating the graph generation process, an innovative approach that could potentially enhance interpretability. 
This will answer the question: what should the road network be like if we know the features? 
This could be achieved by systematically manipulating the GE and subsequently scrutinizing its resultant impact on the graph structure through graph generation techniques. 
In addition, we posit that a comparative analysis of patterns across diverse geographical locations and historical time periods worldwide may shed light on whether changes in the structure of urban road networks resonate with alterations in community structures. 
Such comparative studies have the potential to provide broader insights into the complex interplay of social and infrastructural factors in urban environments.

\section{Acknowledgement}
This material is based upon work supported by the U.S. Department of Energy’s Office of Energy Efficiency and Renewable Energy (EERE) under the Vehicle Technology Program Award Number DE-EE0009211. The views expressed herein do not necessarily represent the views of the U.S. Department of Energy or the United States Government.

\bibliographystyle{elsarticle-harv}
\bibliography{acmart}
\end{document}



\renewcommand{\shortauthors}{XXX and XXX, et al.}






\maketitle

\section{Data Description}

We use seven real-world spatiotemporal datasets to evaluate our model (see Table \ref{tab:data} for an overview). They are:

\textbf{METR-LA}\footnote{\url{https://github.com/liyaguang/DCRNN}} consists of traffic speed information collected by highway loop detectors in Los Angeles \citep{li2018diffusion,wu2020inductive}. We follow the experiment settings of \citet{li2018diffusion} to select 4 months data from Mar 1st 2012 to Aug 30th 2012 with 207 sensors. 

\textbf{NREL}\footnote{\url{https://www.nrel.gov/grid/solar-power-data.html}} contains many energy datasets, and here we choose the Alabama Solar Power Data for Integration Studies \citep{sengupta2018national}. This dataset includes 5-minute solar power records of 137 photovoltaic power plants in 2006. We follow the work of \citet{wu2020inductive} by only keeping the data from 7 am-7 pm everyday in order to attenuate this effect.

\textbf{PeMS-Bay}\footnote{\url{https://github.com/liyaguang/DCRNN}} is also a traffic speed dataset that is collected in Bay Area by Performance Measurement System (PeMS). Same as the work of \citet{li2018diffusion}, we choose 325 sensors from Jan 1st 2017 to May 13th 2017.

\textbf{NOAA}\footnote{\url{https://github.com/MengyangGu/GPPCA}} records the global gridded air and marine temperature anomalies from U.S. National Oceanic and Atmospheric Administration (NOAA) \citep{gu2020generalized}. NOAA contains the monthly data from Jan 1999 to Dec 2018 with $5^{\circ}\times 5^{\circ}$ latitude-longitude resolution. We follow the work of \citet{gu2020generalized} to leave out polar circles and choose 1639 observed grids. With the largest 20 singular values take only 46.6\% portion, the dataset is hard to be considered low-rank.

\textbf{MODIS}\footnote{\url{https://modis.gsfc.nasa.gov/data/}} consists of daytime land surface measured by the Terra platform on the MODIS satellite with 3255 downsampled grids from Jan 1, 2019 to Jan 16, 2021. It is automatically collected by \textit{MODIStsp} package in \textit{R} \citep{busetto2016modistsp}. We follow the work of \citet{heaton2019case} to select a region with similar longitude-latitude bounds as the target (latitude: 30.5$\sim$37.6; longitude: -96.8$\sim$-89.7). Because of the cloud covering the satellites, there are more than 39.6\% entries are missing in the collected data. Moreover, the missing data are generally the continuous spatial area, which is really challenging for models to train.

\textbf{USHCN}\footnote{\url{https://www.ncdc.noaa.gov/ushcn/introduction}} contains monthly precipitation of 1218 locations from 1899 to 2019, which is collected by the U.S. Historical Climatology Network (USHCN) \citep{menne2009us}. As \citet{wu2020inductive} mentioned, USHCN dataset is pretty dispersed, with variance-to-mean ratio exceeds 500. Thus it can help examine the model performance on time series with substantial oscillations.


We summarize these datasets in Table \ref{tab:data}

\begin{table}[!ht]
\small
\footnotesize
    \centering
    \caption{Real-world spatiotemporal datasets description}
    \begin{tabular}{ccccccc}
    \toprule
            & Sensors & Time length & Distance type& Normalization parameter $\sigma$  & Usage & Missing  \\
                   &  & (frequency)& &  of adjacency matrix  &  &  \\
    \midrule
    METR-LA & 207 & 34272 (5-min)& Travelling distance & 557 & SATCN training & 8.11\%\\
    NREL  & 137 & 105120 (5-min) & Haversine distance &14,000  & SATCN training & - \\
    PeMS-Bay  & 325 & 52116 (5-min)& Travelling distance & 1.5   & SATCN training & 0.003\%\\
    NOAA  & 1639 & 240 (1-month)& Haversine distance& 6,000,000   & SATCN training & -\\
    MODIS  & 3225 & 747 (1-day)& Haversine distance & 112,000   & SATCN training & 39.62\%\\
    USHCN & 1218 & 1440 (1-month)& Haversine distance & 100,000  & SATCN training & 3.07\%\\
    \bottomrule
    \end{tabular}
    \label{tab:data}
\end{table}

\section{Implementation Details of SATCN}

\section{Benchmark Settings}
\textbf{kNN} For traffic datasets METR-LA/PeMS-Bay/PeMSD4, the distance is based on the road distance in traffic network. While haversine distance is used in geospatial datasets NREL/NOAA/MODIS/USHCN. For each dataset, we tune the best $k$ according to the lowest kriging RMSE errors. The selection of $k$ for different datasets includes: \textit{METR-LA:} 7; \textit{NREL:} 10; \textit{PeMS-Bay:} 4; \textit{USHCN:} 7; \textit{NOAA:} 2; \textit{MODIS:} 3 and \textit{USHCN:} 7.

\textbf{Okriging} Ordinary kriging is a common used geostatistical method in interpolating values by fitting prior spatial covariance in Gaussian process. We use \textit{Automap} package in \textit{R} to implement the ordinary kriging \citep{krige1951statistical,carley2012automap}. To keep consistent with the construction of adjacency matrix, we fix the variogram kernels as "Gaussian" and automatically learns their parameters by \textit{autoKrige} function. Note that we apply the method at each snapshot. In other words, at each time point, we provide the corresponding values of observed nodes to the algorithm to obtain the interpolated column vector as kriging result. This task fulfills embarrassingly parallel for all time steps.

\textbf{GLTL} Greedy Low-rank Tensor Learning \citep{bahadori2014fast} is designed for both co-kriging and forecasting tasks for multiple variables. There are only one variable for our datasets. We implement this with the \textit{MATLAB} source code\footnote{\url{http://roseyu.com/code.html}} of the authors. We choose the \textit{Orthogonal} algorithm according to its superior performance in \citep{bahadori2014fast}. We set the maximum number of iterations to 1000 and the convergence stopping criteria to $1\times 10^{-10}$. For all the datasets, we use the same Gaussian kernel based adjacency matrices. The Laplacian matrix is calculated by $L=D-W$ where $D$ is a diagonal matrix with $D_{ii} = \sum_{j}W_{ij}$. Same to the implementation in \cite{bahadori2014fast}, we rescale the Laplacian matrix by ${L}=\frac{L}{\max_{ij} L_{ij}}$ as well. An essential parameter is $\mu$ for the weight of Laplacian regularizer, which is tuned by performing grid search from $\{0.05,0.5,5,50,500\}$. We reach convergence before the maximum 1000 iterations in all datasets. The tuned $\mu$ for different datasets are listed in Table \ref{tab:gltl_para}.

\begin{table}[!ht]
    \centering
    \caption{GLTL Parameters for each dataset}
    \begin{tabular}{ccccccc}
    \toprule
    Parameter & METR-LA & NREL & NOAA & MODIS & PeMS-Bay & USHCN \\
    \midrule
$\mu$ & 0.5 & 50 & 5 & 0.5 & 5 & 5  \\
    \bottomrule
    \end{tabular}
    \label{tab:gltl_para}
\end{table}

\textbf{IGNNK} Inductive Graph Neural Network for Kriging \citep{wu2020inductive} is a novel GNN based model that combines dynamic subgraph sampling techniques and diffusion graph convolution structure for the kriging task. We implement this from the Github repository\footnote{\url{https://github.com/Kaimaoge/IGNNK}} of \citep{wu2020inductive}. For datasets with distance information, it uses a Gaussian kernel to construct the adjacency matrix for GNN:
\begin{equation}
W_{ij} = \exp\left(-\left(\frac{\text{dist}\left(v_i , v_j\right)}{\sigma}\right)^2\right),
\label{distance_rule}
\end{equation}
where $W_{ij}$ stands for adjacency or closeness between sensors/nodes $v_i$ and $v_j$, $\text{dist}\left(v_i , v_j\right)$ is the distance between $v_i$ and $v_j$, and $\sigma$ is the normalization parameter, which is illustrated in Table \ref{tab:data}. We use Adam optimizer with learning rate 0.0001 to optimize the GNN training and set the maximum number of training episodes as 750. Apart from them, there are many identical model parameters defined for each dataset. We follow similar settings in the work \citet{wu2020inductive}, which are listed in Table \ref{tab:para}:

\begin{table}[!ht]
\small
    \centering
    \caption{IGNNK Parameters for each dataset}
    \begin{tabular}{ccccccc}
    \toprule
{Parameters} & METR-LA & NREL & USHCN & NOAA & PeMS-Bay & MODIS \\
    \midrule
window length $h$ & 24 & 16 & 6 & 1 & 24 & - \\
number of evaluation windows (test) & 428 & 1971 & 72 & 72 & 2171 & - \\
number of kriging nodes ($n_t^u$) & 50 & 30 & 300 & 500 & 80 & - \\
sampled observed nodes size $n_o$ & 100 & 100 & 900 & 1100 & 240 & - \\
sampled masked nodes size $n_m$ & 50 & 30 & 300 & 400 & 80 & - \\
hidden feature dimension $z$ &100 & 100 & 100 & 100 & 100 & - \\
activation function $\sigma$ & \textit{relu} & \textit{relu} & \textit{relu} & \textit{relu} & \textit{relu} & - \\
batch size ($S$)  & 4 & 8 & 8 & 2 & 4 & - \\
number of iterations ($I_{\max}$) & 186750 & 287250 & 30750 & 63000 & 285000 & - \\
order of diffusion convolution & 2 & 2 & 2 & 2 & 2 & - & 2\\ 
    \bottomrule
    \end{tabular}
    \label{tab:para}
\end{table}

\textbf{KCN-Sage} We implemented a Pytorch version KCN following the code\footnote{https://github.com/tufts-ml/KCN} given by the original authors. The authors implemented several graph neural networks based kriging models, including GCN, GAT, and GraphSage. According to the experimental results, the KCN-Sage model achieves the best performance. So we compare SATCN with KCN-Sage. KCNs are originally proposed for kriging task under a fixed graph structure, we use Algorithm 1 to train them adpative to our task. As the same as \citep{appleby2020kriging}, a 4-layer GraphSage with hidden size [20, 10, 5, 3] is used. The values of 3 nearest neighbors are used as inputs for KCN-Sage.  
\bibliographystyle{ACM-Reference-Format}
\bibliography{acmart}